\documentclass[times,5p,twocolumn,lefttitle,final,longtitle]{elsarticle}
\usepackage[utf8]{inputenc}
\usepackage{newunicodechar}
\newunicodechar{−}{-}

\usepackage{graphicx} 
\usepackage{booktabs}
\usepackage{amsmath}
\usepackage{url}
\usepackage{pdflscape}
\usepackage{booktabs}
\usepackage[table,xcdraw]{xcolor}
\usepackage{dblfloatfix}
\usepackage{tikz}
\usepackage{lipsum}
\usepackage{float}
\usepackage[colorlinks]{hyperref}
\usepackage[nolist]{acronym}
\usepackage{xspace}
\usepackage{caption}

\usepackage{etoolbox}
\AtBeginEnvironment{thebibliography}{%
  \footnotesize
  \setlength{\bibsep}{0.3ex}%
  \setlength{\itemsep}{0pt}%
  \setlength{\parsep}{0pt}%
  \setlength{\topsep}{0pt}%
}

\definecolor{bcol}{HTML}{377EA8}
\AtBeginDocument{%
\hypersetup{
citecolor=bcol,
linkcolor=bcol,
urlcolor=bcol
}
}
\definecolor{uktfdg}{HTML}{176D9C}
\definecolor{lmufdg}{HTML}{BA8A39}
\definecolor{uktpsma}{HTML}{40896C}
\definecolor{lmupsma}{HTML}{AE662E}

\newcommand{\lmupsma}{PSMA\textsubscript{LMU}\xspace}
\newcommand{\lmufdg}{FDG\textsubscript{LMU}\xspace}
\newcommand{\uktpsma}{PSMA\textsubscript{UKT}\xspace}
\newcommand{\uktfdg}{FDG\textsubscript{UKT}\xspace}

\newcommand{\ap}{autoPET3\xspace}

\newcommand{\teamcolor}[1]{
\begin{tikzpicture}
\draw [fill=#1, draw=#1]circle (1.0mm);
\end{tikzpicture}}

\biboptions{authoryear}    

\title{The \ap Challenge: Automated Lesion Segmentation in Whole-Body PET/CT -- Multitracer Multicenter Generalization}

\cortext[cor1]{Corresponding author:~jakob.dexl@lmu.de} 
\author[1,2]{Jakob Dexl\corref{cor1}}
\ead{jakob.dexl@lmu.de}

\author[1,2,17]{Katharina Jeblick}
\author[1,2]{Andreas Mittermeier}
\author[1,2]{Balthasar Schachtner}
\author[1,2]{Anna Theresa Stüber}
\author[1,2,18]{Johanna Topalis}

\author[5,8]{Maximilian Rokuss}
\author[5,6]{Fabian Isensee}
\author[5,6,7]{Klaus H. Maier-Hein}

\author[9,10]{Hamza Kalisch}
\author[9,15]{Jens  Kleesiek}
\author[9,10]{Constantin M. Seibold}

\author[11]{Hussain Alasmawi}

\author[12]{Lap Yan Lennon Chan}
\author[13]{Yixuan Yuan}

\author[14,15]{Alexander Jaus}
\author[14,15]{Rainer Stiefelhagen}

\author[3]{Pauline Ornela Megne Choudja}
\author[3,19]{Konstantin Nikolaou}
\author[3]{Christian La Fougère}
\author[4]{Sergios Gatidis}

\author[1]{Matthias P. Fabritius}
\author[1]{Maurice Heimer}
\author[1]{Gizem Abaci}

\author[1]{Lalith Kumar Shiyam Sundar}

\author[16]{Rudolf A. Werner}

\author[1]{Jens Ricke}
\author[1]{Clemens C. Cyran}

\author[3]{Thomas Küstner}
\ead{thomas.kuestner@med.uni-tuebingen.de}

\author[1,2,18]{Michael Ingrisch}
\ead[url]{michael.ingrisch@med.uni-muenchen.de}

\affiliation[1]{organization={Department of Radiology, LMU University Hospital, LMU Munich},
city={Munich},
country={Germany}}

\affiliation[2]{organization={Munich Center for Machine Learning (MCML)},
city={Munich},
country={Germany}}

\affiliation[3]{organization={University Hospital Tübingen, Department of Radiology},
city={Tübingen},
country={Germany}}

\affiliation[4]{organization={Department of Radiology, Stanford University},
city={Stanford},
country={USA}}

\affiliation[5]{organization={German Cancer Research Center (DKFZ)},
city={Heidelberg},
country={Germany}}

\affiliation[6]{organization={Helmholtz Imaging, DKFZ},
city={Heidelberg},
country={Germany}}

\affiliation[7]{organization={Pattern Analysis and Learning Group, Department of Radiation Oncology,
Heidelberg University Hospital},
city={Heidelberg},
country={Germany}}

\affiliation[8]{organization={Faculty of Mathematics and Computer Science, Heidelberg University},
city={Heidelberg},
country={Germany}}

\affiliation[9]{organization={Institute for AI in Medicine (IKIM), University Hospital Essen (AöR)},
city={Essen},
country={Germany}}

\affiliation[10]{organization={Department of Nuclear Medicine, University Hospital Essen (AöR)},
city={Essen},
country={Germany}}

\affiliation[11]{organization={Mohamed bin Zayed University of Artificial Intelligence},
city={Abu Dhabi},
country={UAE}}

\affiliation[12]{organization={Department of Computer Science and Engineering, The Chinese University of Hong
Kong},
city={Shatin},
country={Hong Kong SAR}}

\affiliation[13]{organization={Department of Electronic Engineering, The Chinese University of Hong Kong},
city={Shatin},
country={Hong Kong SAR}}

\affiliation[14]{organization={Karlsruhe Institute of Technology},
city={Karlsruhe},
country={Germany}}

\affiliation[15]{organization={HIDSS4Health - Helmholtz Information and Data Science School for Health},
city={Karlsruhe/Heidelberg},
country={Germany}}

\affiliation[16]{organization={Department of Nuclear Medicine, LMU University Hospital, LMU Munich},
city={Munich},
country={Germany}}

\affiliation[17]{organization={Comprehensive Pneumology Center (CPC-M), Member of the German Center for Lung Research (DZL)},
city={Munich},
country={Germany}}

\affiliation[18]{organization={relAI – Konrad Zuse School of Excellence in Reliable AI},
city={Munich},
country={Germany}}

\affiliation[19]{organization={Cluster of Excellence iFIT (EXC 2180) "Image Guided and Functionally Instructed Tumor Therapies", University of Tübingen},
city={Tuebingen},
country={Germany}}

\journal{Medical Image Analysis}
\begin{document}

\begin{acronym}
  \acro{ac1}[AC1]{Award Category 1}
  \acro{ac2}[AC2]{Award Category 2}
  \acro{dsc}[DSC]{Dice Similarity Score}
  \acro{fpv}[FPV]{False Positive Volume}
  \acro{fnv}[FNV]{False Negative Volume}
  \acro{psma}[PSMA]{Prostate-Specific Membrane Antigen}
  \acro{fdg}[FDG]{Fluorodeoxyglucose}
\end{acronym}

\begin{abstract}
    We report the design and results of the third autoPET challenge (MICCAI 2024), which benchmarked automated lesion segmentation in whole-body PET/CT under a compositional generalization setting. Training data comprised 1,014 [\textsuperscript{18}F]-FDG PET/CT studies from the University Hospital Tübingen and 597 [\textsuperscript{18}F]/[\textsuperscript{68}Ga]-PSMA PET/CT studies from the LMU University Hospital Munich, constituting the largest publicly available annotated PSMA PET/CT dataset to date. The held-out test set of 200 studies covered four tracer--center combinations, two of which represented unseen compositional pairings. A complementary data-centric award category isolated the contribution of data handling strategies by restricting participants to a fixed baseline model. Seventeen teams submitted 27 algorithms, predominantly nnU-Net-based 3D networks with PET/CT channel concatenation. The top-ranked algorithm achieved a mean DSC of 0.66, FNV of 3.18 mL, and FPV of 2.78 mL across all four test conditions, improving DSC by 8\% and reducing the false-negative volume by 5\,mL relative to the provided baseline. Ranking was stable across bootstrap resampling and alternative ranking schemes for the top tier. Beyond the benchmark, we provide an in-depth analysis of segmentation performance at the patient and lesion level. Three main conclusions can be drawn: (1) in-domain multitracer PET/CT segmentation is sufficient and probably approaching reader agreement; (2) compositional generalization to unseen tracer--center combinations remains an open problem mainly driven by systematic volume overestimation; (3) heterogeneity and case difficulty drive performance variation substantially more than the choice of algorithm among top-ranked teams. 

\end{abstract}

\maketitle

\section{Introduction}
Positron Emission Tomography/Computed Tomography (PET/CT) integrates molecular and anatomical information within a single examination \citep{beyer_combined_2000, townsend_multimodality_2008} and has become an established imaging modality across multiple clinical domains, including oncology \citep{rohren_clinical_2004, boellaard_international_2024}, cardiology \citep{slart_total-body_2024}, and neurology \citep{xie_pet_2024}. Among these, oncologic imaging represents the most widespread application, where PET/CT is routinely used for diagnosis, staging, radiotherapy planning, and treatment response assessment, thereby substantially influencing patient management across a wide range of tumor entities. With rising cancer incidence worldwide \citep{bray_global_2024} and aging populations, both the volume of PET/CT examinations and the demand for reliable interpretation are expected to increase considerably.

[\textsuperscript{18}F]-\ac{fdg} remains the most widely used PET radiotracer, demonstrating high sensitivity in glucose-avid malignancies such as lymphomas \citep{cheson_recommendations_2014}, melanoma \citep{boellaard_fdg_2015}, and non-small-cell lung cancer \citep{ettinger_nccn_2023}. However, it performs poorly in tumors with low glycolytic activity (including well-differentiated neuroendocrine tumors, hepatocellular carcinoma, and prostate cancer), driving the development of target-specific tracers, among which \ac{psma}-targeted agents ([\textsuperscript{68}Ga] or [\textsuperscript{18}F]) have emerged as highly effective radiotracers for prostate cancer imaging \citep{fendler_assessment_2019}.

Despite advances in molecular imaging, routine PET/CT interpretation remains largely visual or semi-quantitative. Whole-body quantitative metrics, including total metabolic tumor volume (TMTV) and total lesion glycolysis \citep{larson_tumor_1999}, provide complementary prognostic and treatment-stratification information across multiple malignancies \citep{sasanelli_pretherapy_2014, meignan_total_2021, mikhaeel_proposed_2022, seifert_prognostic_2023, pak_prognostic_2014, im_prognostic_2015}. Yet their adoption is limited by the time-consuming and labor-intensive nature of manual or semi-manual segmentation. Current image interpretation time ranges from five to 90 minutes per patient, with typical readings taking around 30 minutes \citep{ratib_petct_2004, brady_guidelines_2025, beyer_variations_2011}, making it untenable to allocate additional time for manual segmentation. Automated tumor segmentation overcomes these limitations by enabling systematic extraction of these imaging biomarkers, providing objective, reproducible measures of tumor burden, distribution, and spatial heterogeneity to support standardized reporting, improve patient management, and facilitate downstream research.

\begin{figure}[t]
\centering
\includegraphics[width=\columnwidth]{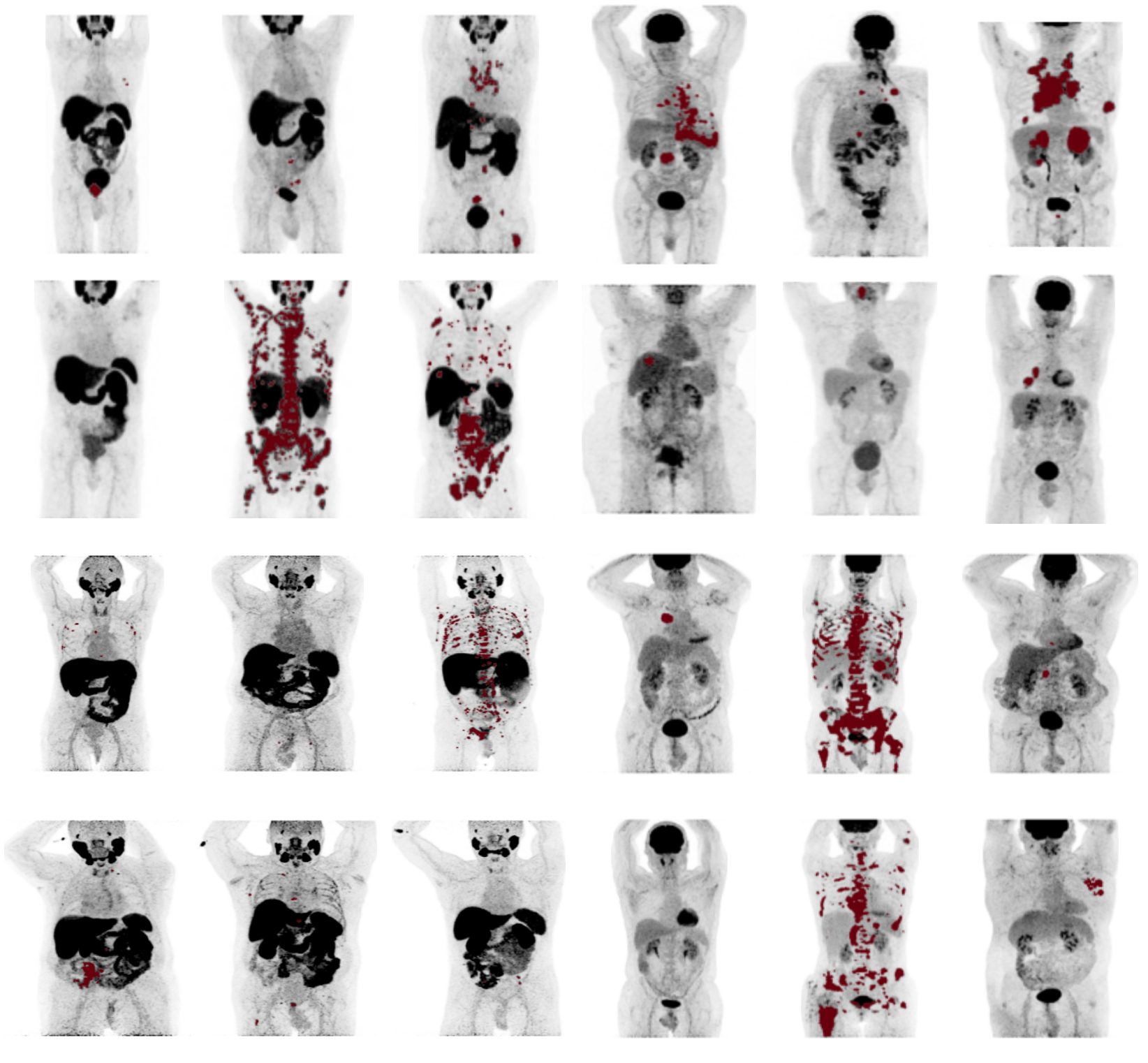}
\caption{Representative test cases illustrating the key challenges for automated lesion segmentation: substantial inter-patient heterogeneity in disease extent, from a single lesion to a high metastatic burden, as well as tracer-specific physiological uptake patterns. Cases include PSMA (left) and FDG (right) scans from the LMU (top) and UKT (bottom) cohorts. Red overlays denote manual segmentations; intensity reflects tracer uptake.}\label{fig:fig1}
\end{figure} 

From a computational image analysis perspective, the core challenge is distinguishing pathological from physiological uptake against a tracer-specific background. For \ac{fdg}, physiological activity is dominated by the brain, myocardium, and urinary tract, whereas \ac{psma}-targeted tracers show high physiological uptake in the lacrimal and salivary glands, liver, kidneys, bowel, ureters and urinary bladder. This is further complicated by numerous pitfalls that challenge even experienced readers. For \ac{fdg}, common examples include things like brown fat activation, post-exercise muscle uptake, or injection-site artifacts such as tracer extravasation which can mimic malignancy, while small or low \ac{fdg} avidity lesions may go undetected \citep{rosenbaum_false-positive_2006, simpson_fdg_2017}. For \ac{psma}, notable false positives stem from ganglionic uptake mimicking lymph node metastases, post-radiotherapy remodeling, inflammation, and false negatives from low \ac{psma} expression, small lesion size or adjacent uptake in the urinary bladder \citep{sheikhbahaei_pearls_2017,fendler_assessment_2019, fendler_false_2021}. In addition, the acquired images may be affected by CT-based attenuation correction artifacts \citep{blodgett_petct_2011}, respiratory motion causing PET/CT misregistration and SUV underestimation \citep{nehmeh_respiratory_2008}, and partial-volume effects in small lesions \citep{soret_partial-volume_2007}. Finally, as shown in Figure \ref{fig:fig1}, cases exhibit significant variation in disease extent, from a single lesion to a high metastatic burden, posing a fundamental challenge for automated segmentation.

An additional challenge commonly reported is limited robustness under distribution shifts, for example, when deploying segmentation models to data from different centers. In PET/CT challenges, segmentation performance has consistently degraded on data from held-out institutions \citep{oreiller_head_2022, gatidis_results_2024, dexl_autopet_2025}. These shifts are typically composites, involving differences in scanner hardware, reconstruction protocols, patient populations, disease characteristics, and annotation practices, which remain difficult to disentangle. Furthermore, even SUV measurements exhibit substantial inter-scanner and inter-protocol variability, as documented in systematic reviews \citep{adams_systematic_2010} and multi-site phantom studies \citep{fahey_variability_2010}.

The autoPET challenge series \citep{gatidis_results_2024, dexl_autopet_2025} aims to advance automated PET/CT image analysis by establishing a transparent benchmark and providing large, machine-learning-ready datasets for algorithm development and evaluation. Building on previous editions, \ap is the first challenge to explicitly address multitracer multicenter generalization in automated lesion segmentation. To this end, we release the largest publicly available annotated \ac{psma} PET/CT dataset to date, comprising 597 scans of prostate cancer patients, which complements the \ac{fdg} PET/CT datasets released in previous editions. The resulting combination of tracers and institutions introduces domain shifts arising from differences in tracer biodistribution, scanner hardware, reconstruction protocols, and patient populations. Algorithms are therefore evaluated in a novel compositional setting, in which tracer type and acquisition site are entangled in the training data and models must generalize to unseen combinations of both factors. In addition to the primary segmentation task, \ap introduces the first data-centric track in a PET/CT challenge, in which participants aim to improve upon a fixed baseline model exclusively through pre-processing, augmentation, and training pipelines. This track is motivated by the observation that data handling substantially influenced performance in previous challenge editions and remains insufficiently explored in PET image analysis. A systematic post-challenge analysis further examines segmentation performance at the patient and lesion level.

\section{Related Works}
\subsection{Related medical image segmentation challenges}
Multiple PET and PET/CT challenges have been held over the past years, differing in modality, dataset size, target entity, and field of view (FOV). The first MICCAI challenge on PET tumor segmentation \citep{hatt_first_2018} used a small dataset of phantom, synthetic, and clinical images of isolated solid tumors, with most methods relying on classical machine learning. It established the value of benchmarking for advancing PET segmentation but remained limited in scale and clinical representativeness.

The Head and Neck Tumor (HECKTOR) challenge series focused on PET/CT-based segmentation and outcome prediction in head-and-neck cancer patients. Over three editions (2020–2022) \citep{oreiller_head_2022, andrearczyk_overview_2023, andrearczyk_automatic_2023}, the challenge progressively expanded in scale and scope, culminating in multi-class segmentation of primary and nodal tumor volumes on 883 cases from nine institutions, alongside survival and recurrence prediction tasks. U-Net-based architectures consistently dominated the leaderboard, and performance improved with increasing dataset size and diversity, while segmentation of nodal disease and generalization across centers remained challenging. Across editions, segmentation quality was strongly dependent on tumor size and metabolic activity, and by the final edition, results were considered potentially sufficient for clinical use. 

The automated lesion segmentation in whole-body PET/CT (autoPET, 2022) series focuses on whole-body tumor segmentation in lung cancer, lymphoma, and melanoma patients. The first iteration \citep{gatidis_results_2024} served primarily as a proof of concept, providing one of the largest, publicly accessible datasets for whole-body \ac{fdg}-PET/CT lesion segmentation with expert annotations. The training set consists of 1,014 studies (900 patients) acquired at a single site. The held-out test set for final evaluation comprised 150 PET/CT studies: 100 from the same hospital as the training data and 50 from a different hospital, to assess generalizability across centers. Participants submitted mainly 3D U-Nets, which demonstrated the feasibility of accurate lesion segmentation under those conditions, while noting that algorithm performance remained dependent on data quantity, data quality, and design choices such as post-processing. 

The second iteration (autoPET2, 2023) \citep{dexl_autopet_2025} extended the scope of the challenge by focusing on single-source domain generalization. Models trained on the same source distribution were evaluated across multiple clinically distinct target domains, assessing robustness to variations in scanner hardware, patient demographics, pathology, and a different PET tracer (\ac{psma}). The results highlighted the limitations of current methods when confronted with out-of-distribution data and underscored the need for more diverse training datasets and improved strategies to achieve reliable real-world deployment. The highest-ranked submission achieved an average \ac{dsc} slightly above 0.50, with notable performance degradation on out-of-domain pediatric and \ac{psma} tracer data, where false positives from physiological uptake and decreased sensitivity for small or low-uptake lesions remained frequent failure modes.

\subsection{Related tumor segmentation algorithms}
\label{sec:relatedalgo}
Parallel to challenge-driven efforts, the broader literature on PET tumor segmentation has shifted from classical rule-based or semi-automatic methods \citep{foster_review_2014} towards fully learning-based approaches, mainly due to the availability of large annotated PET/CT datasets.

\ac{fdg} PET/CT has been the primary focus of deep learning–based tumor detection and segmentation research. Early approaches formulated the task as the classification of predefined \ac{fdg}-avid candidate regions. \citet{sibille_18f-fdg_2020} used SUV-based thresholding to extract candidate foci and trained a CNN to classify them as malignant or benign in 629 lung cancer and lymphoma patients, achieving an AUC of 0.98. More recent work has targeted end-to-end detection and segmentation, using architectures such as patch-wise CNNs in 90 lymphoma patients \citep{weisman_convolutional_2020}, Retina U-Net for lung cancer staging in 364 patients \citep{weikert_automated_2023}, a two-stage U-Net for lung tumor delineation in 887 patients (\ac{dsc} 0.78) \citep{park_automatic_2023}, and 3D U-Net variants for lymphoma segmentation and TMTV estimation in cohorts of 733 (\ac{dsc} 0.73) \citep{blanc-durand_fully_2021} and 1418 patients (\ac{dsc} 0.68) \citep{yousefirizi_tmtv-net_2024}. Across these studies, common failure modes persist. False positives are predominantly driven by physiological uptake in bone marrow, mediastinal structures \citep{weikert_automated_2023}, and brown adipose tissue \citep{weisman_convolutional_2020}. False negatives consistently involve small or low uptake lesions \citep{weikert_automated_2023, weisman_convolutional_2020, park_automatic_2023}. External validation remains limited: \citet{weikert_automated_2023} confirmed similar performance on 20 external cases, and \citet{yousefirizi_tmtv-net_2024}, which was trained on the autoPET \ac{fdg} data, demonstrated only a 2\% \ac{dsc} drop across 518 multi-center scans, whereas \citet{blanc-durand_fully_2021} observed a 5\% \ac{dsc} drop and a significant TMTV underestimation of 20.8\% on their external cohort. 

Methods for segmenting PSMA PET/CT, on the other hand, are scarce and more recent. \citet{kendrick_fully_2022} employed a 3D nnU-Net cascade on 337 [\textsuperscript{68}Ga]Ga-\ac{psma}-11 scans from a single center, reporting a mean voxel-level \ac{dsc} of 0.44, notably within measured inter-observer variability. \citet{jafari_convolutional_2024} applied the same nnU-Net cascade framework to 412 multicenter [\textsuperscript{68}Ga]Ga-\ac{psma}-11 scans, achieving an internal \ac{dsc} of 0.70 with moderate drops on two external test sets (0.65 and 0.68). \citet{yazdani_automated_2024} used 652 unlabeled [\textsuperscript{68}Ga]Ga-\ac{psma}-11 scans for self-supervised pretraining of a Swin-UNETR encoder and fine-tuned on 100 labeled cases, reporting a \ac{dsc} of 0.68. More recently, \citet{leung_deep_2024} demonstrated a semi-supervised transfer-learning strategy that jointly leveraged 611 \ac{fdg} and 408 \ac{psma} PET/CT scans with incomplete annotations, achieving a median \ac{dsc} of 0.73 for prostate cancer and setting a useful precedent for cross-tracer domain adaptation. Across all these studies, image data remain institutional and labeling is often partially incomplete, underscoring the need for a public whole-body \ac{psma} benchmark.

\section{Material and Methods}

\subsection{Challenge Task}
The \ap challenge was designed to evaluate automated lesion segmentation in whole-body PET/CT scans across two tracers and institutions in a compositional generalization setting. To capture both algorithmic flexibility and the effects of data-centric strategies, the challenge defined two complementary award categories:

\textbf{\ac{ac1} -- Best generalizing model:}  
Teams could freely choose any model architecture, ensemble strategy, or training configuration. Publicly available external datasets and pretrained models were permitted if appropriately cited. The focus was on developing the most generalizable segmentation model across heterogeneous imaging domains.

\textbf{\ac{ac2} -- Data-centric excellence:}  
Participants were restricted to using only the provided training data and a fixed baseline model architecture\footnote{\url{https://github.com/ClinicalDataScience/datacentric-challenge}}. Modifications were limited to the data pipeline (e.g., augmentation, error handling, synthetic data generation) to assess the benefits of pre-processing choices.

\subsection{Challenge Organization and Infrastructure}
The challenge was organized in conjunction with the 27th International Conference on Medical Image Computing and Computer-Assisted Intervention (MICCAI) 2024 and coordinated by a multidisciplinary team from the Ludwig Maximilian University Hospital Munich (LMU) and the University Hospital Tübingen (UKT), Germany. The proposal was peer-reviewed and accepted by the MICCAI challenge committee in early 2024\footnote{\url{https://zenodo.org/records/10990932}}.

All challenge materials, including rules, dataset descriptions, baseline implementations, and evaluation details, were made publicly available via the Grand Challenge platform\footnote{\url{https://autopet-iii.grand-challenge.org}} and an accompanying GitHub repository\footnote{\url{https://github.com/ClinicalDataScience/autoPETIII}}. The training dataset was released on April 1, 2024, under the CC-BY-NC 4.0 license.

Submissions were evaluated by running Docker containers in a standardized, secure environment on held-out test data. Each submission was limited to five minutes of runtime per case and executed without access to auxiliary metadata. The evaluation infrastructure consisted of NVIDIA T4 GPUs (16 GB VRAM), 8 CPU cores, and 32 GB RAM. A preliminary test phase enabled technical validation, followed by a final test phase in September 2024. Results were presented at an in-person session during MICCAI 2024 and subsequently released online. The top three teams in \ac{ac1} and the top two teams in \ac{ac2} were invited to present their methods.

\begin{table*}[!t]
\centering
\def\arraystretch{1.1}
\caption{Overview of the datasets used for training and testing across different centers and tracers. The colors for the test data are used throughout the paper in all figures. Data are represented as number or mean $\pm$ SD.}
\label{tab:dataset}
\small
\begin{tabular}{@{}p{2cm}*{6}{p{2cm}}@{}}
\toprule
Name          & FDG\textsubscript{train}    & PSMA\textsubscript{train} & \teamcolor{uktfdg} \uktfdg  & \teamcolor{lmupsma} \lmupsma & \teamcolor{lmufdg} \lmufdg & \teamcolor{uktpsma} \uktpsma \\ \midrule
Split         & Train     & Train   & Test      & Test      & Test     & Test      \\
Center        & Tuebingen & Munich  & Tuebingen & Munich    & Munich   & Tuebingen    \\
Tracer &
  {[}\textsuperscript{18}F{]}FDG &
\begin{tabular}[t]{@{}l@{}}
  {[}\textsuperscript{18}F{]}PSMA-1007\\
  {[}\textsuperscript{68}Ga{]}PSMA-11
\end{tabular} &
{[}\textsuperscript{18}F{]}FDG &
\begin{tabular}[t]{@{}l@{}}
  {[}\textsuperscript{18}F{]}PSMA-1007\\
  {[}\textsuperscript{68}Ga{]}PSMA-11
\end{tabular} &
{[}\textsuperscript{18}F{]}FDG &
{[}\textsuperscript{18}F{]}PSMA-1007 \\
  N Scanner     & 1         & 3       & 1         & 3         & 3        & 1         \\ 
Resolution (mm\textsuperscript{3})     &  2.04\textsuperscript{2} \texttimes~3.00        &\begin{tabular}[t]{@{}l@{}}
  2.73\textsuperscript{2} \texttimes~3.27\\
  4.07\textsuperscript{2} \texttimes~2.00\\
  4.07\textsuperscript{2} \texttimes~5.00
\end{tabular}       & 2.04\textsuperscript{2} \texttimes~3.00         & \begin{tabular}[t]{@{}l@{}}
  2.73\textsuperscript{2} \texttimes~3.27\\
  4.07\textsuperscript{2} \texttimes~2.00\\
  4.07\textsuperscript{2} \texttimes~5.00
\end{tabular}         & \begin{tabular}[t]{@{}l@{}}
  2.73\textsuperscript{2} \texttimes~3.27\\
  4.07\textsuperscript{2} \texttimes~2.00\\
  4.07\textsuperscript{2} \texttimes~5.00 \end{tabular}        &  2.04\textsuperscript{2} \texttimes~3.00         \\
    \hline
Patients      & 900       & 378     & 50        & 50        & 50       & 50          \\
N Studies     & 1,014      & 597     & 50        & 50        & 50       & 50        \\
Sex (w/m)     & 444/570   & 0/597   & 22/28     & 0/50      & 21/29    & 0/50        \\
\begin{tabular}[t]{@{}l@{}}Age \end{tabular} &
  60 ± 16 &
  71 ± 8 &
  55 ± 12 &
  70 ± 8 &
  53 ± 19 &
  70 ± 8 \\
\begin{tabular}[t]{@{}l@{}}Weight {(kg)}\end{tabular} &
  79 ± 19 &
  83 ± 14 &
  92 ± 21 &
  84 ± 13 &
  73 ± 16 &
  87 ± 16 \\
\hline
\begin{tabular}[t]{@{}l@{}}N Studies \\ w/ lesions\end{tabular} &
  501 &
  537 &
  31 &
  33 &
  50 &
  42 \\
N lesions     & 8,781      & 19,377   & 516       & 1,309      & 410      & 624      \\
TMTV {(mL)} & 110,189    & 132,853  & 7,127      & 6,052      & 5,701     & 882      \\ 
\bottomrule
\end{tabular}%
\end{table*}

\subsection{Participation policies}
The challenge was open to all registered participants with a grand challenge account. Members of the organizing institutions were excluded from eligibility for awards to avoid conflicts of interest. 
Participants could submit up to two algorithms, distributed across the two award categories at their discretion. \ac{ac1} imposed no restrictions on methodology. \ac{ac2} was designed as a data-centric challenge with stricter constraints. No additional data was permitted, and the provided reference network had to be used unchanged. AI models could be employed in the data pipeline (e.g., for augmentation or synthetic data generation), but were not allowed to produce the final prediction. Pretrained models were permitted only if publicly available and not trained on any data beyond the challenge dataset.

All teams were required to publicly release their code and trained model weights under a permissive license and to submit a technical report or preprint describing their approach. No embargo was imposed on post-challenge publications. Although prize funding was initially planned, it could not be awarded due to funding limitations.

\subsection{Challenge Datasets}
\begin{figure}
\centering
\includegraphics[width=\columnwidth]{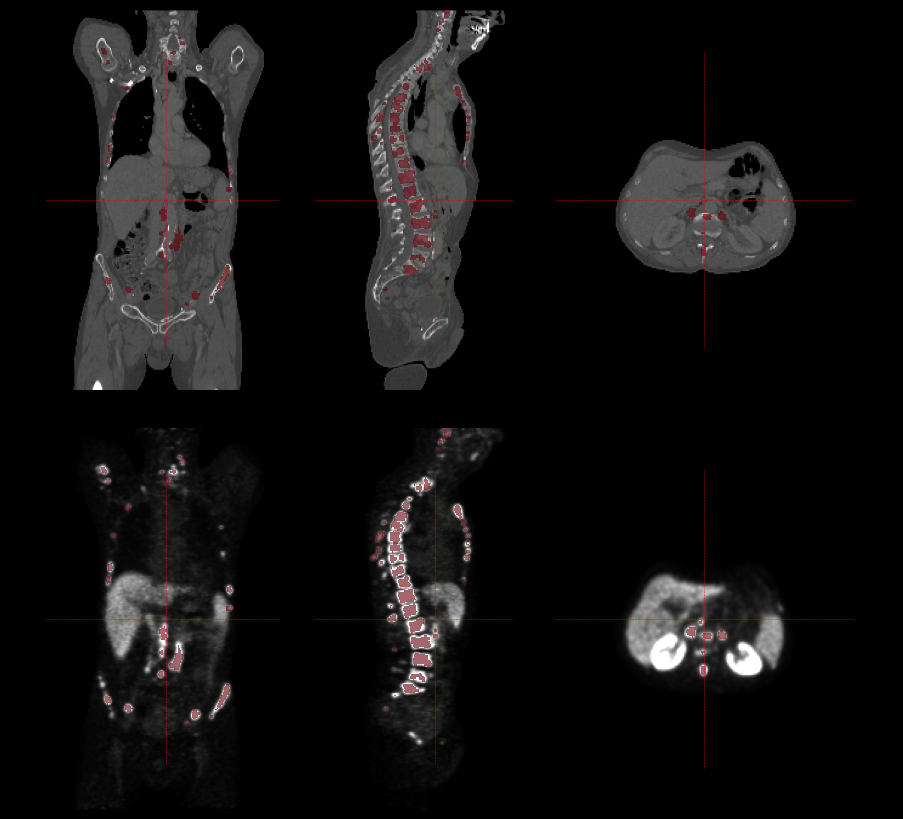}
\caption{Representative \lmupsma PET/CT case shown in three orthogonal planes (coronal, sagittal, axial). Top row: CT images displayed with a window of [400. 1800] Hounsfield units. Bottom row: corresponding PET images displayed as SUV with a window of [0, 10]. Red overlays indicate manual lesion annotations.}\label{fig:caseexample}
\end{figure}

An overview of all datasets is provided in Table~\ref{tab:dataset}. A representative case is shown in Figure \ref{fig:caseexample}. Training data comprises two publicly available whole-body PET/CT datasets that differ in tracer, scanner hardware, and acquisition protocol. Publication of the anonymized datasets was approved by the respective institutional ethics committees and data privacy review boards.
 
The first training set \citep{gatidis_whole-body_2022} contains 1,014 [$^{18}$F]\ac{fdg} PET/CT studies from 900 oncologic patients, acquired at a single institution (UKT) on a Siemens Biograph mCT 120 following a standardized protocol: $\geq$6h fasting, weight-adapted injected activity (mean 314.7\,MBq, SD 22.1\,MBq, range 150--432\,MBq), and 60\,min uptake time. This dataset was released as part of the first autoPET challenge via The Cancer Imaging Archive\footnote{\url{https://www.cancerimagingarchive.net/collection/fdg-pet-ct-lesions/}}. 
 
The second training set \citep{jeblick2025psma} includes 597 \ac{psma} PET/CT examinations from 378 prostate-carcinoma patients (LMU), acquired with two tracers ([$^{18}$F]\ac{psma}-1007 and [$^{68}$Ga]\ac{psma}-11) on three scanners (Siemens Biograph mCT Flow 20, Siemens Biograph 64-4R TruePoint, GE Discovery 690). In contrast to the standardized \ac{fdg} protocol, acquisition parameters varied substantially: injected activities were $246\pm27$\,MBq ([$^{18}$F]) and $214\pm45$\,MBq ([$^{68}$Ga]); uptake times showed pronounced heterogeneity ($74\pm22$\,min, range 6--181\,min and $74\pm19$\,min, range 7--195\,min, respectively). DICOM data are available via The Cancer Imaging Archive\footnote{\url{https://www.cancerimagingarchive.net/collection/psma-pet-ct-lesions/}}; due to extended head-region anonymization required by the platform, the exact challenge data is provided in a separate repository\footnote{\url{https://fdat.uni-tuebingen.de/records/0zs4c-89f12}}.
 
The held-out test set comprises 200 studies. Half originate from the same center and tracer combinations as the training data (50 \ac{fdg} from UKT, 50 \ac{psma} from LMU); the other half constitutes a cross-center evaluation (50 \ac{psma} from UKT, 50 \ac{fdg} from LMU), explicitly assessing compositional generalization across tracers and institutions.
 
All PET images were converted to standardized uptake values (SUV) via $\mathrm{SUV} = C_{\mathrm{tissue}}/(A_{\mathrm{inj}}/ W)$, where $C_{\mathrm{tissue}}$ is the tissue radioactivity concentration (MBq/ml), $A_{\mathrm{inj}}$ the decay-corrected injected activity (MBq), and $W$ the patient body weight (g).

\subsection{Annotation procedure}
Lesion annotation for the \ac{psma} dataset was performed by a medical imaging specialist (S.G., with 3 years of experience in hybrid imaging) and independently verified by two board-certified experts with four and 10+ years of experience, respectively. All annotations were conducted using CE-certified software (Mint Lesion™, Mint Medical, Heidelberg, Germany). SUVs were analyzed alongside the corresponding CT images, either in parallel view or as fused overlays. Regions with elevated \ac{psma} uptake were delineated in 3D by defining circular volumes of interest. Voxels with SUV values above a user-defined threshold were automatically pre-segmented and subsequently manually refined slice-by-slice to generate 3D binary segmentation masks. The SUV threshold was adjusted individually for each case based on visual assessment and used solely to accelerate manual delineation.

\paragraph{Critique and justification}
Threshold-based segmentation approaches, such as those using fixed or relative percentages of the maximum SUV, offer good reproducibility across readers. The European Association of Nuclear Medicine guidelines \citep{boellaard_fdg_2015} specifically recommend 3D isocontours at 41\% and 50\% of the maximum voxel value for reporting metabolic tumor volume and total lesion glycolysis, to improve reproducibility.
However, it is also noted that these VOIs may be unreliable in lesions with heterogeneous uptake, low tumor-to-background contrast, or regions of high uptake nearby, and expert review or adjustment is advised to ensure accurate delineation. Similarly, \citet{hatt_classification_2017} emphasizes that simple threshold-based approaches can underperform in realistic imaging conditions, and should always be verified and, if needed, adapted by a physician for quantitative or radiomic analyses.
Between the \uktfdg data and the other three datasets is an annotation shift. While the \uktfdg data \citep{gatidis_whole-body_2022} is segmented slice-by-slice, the others are based on a threshold algorithm with manual adjustments.   

\subsection{Assessment Methods}
\subsubsection{Metrics}
Segmentation algorithms were evaluated using the same three measures established in the previous autoPET editions to ensure consistency and comparability across the challenges. These metrics included the overall \ac{dsc}, \ac{fpv}, and \ac{fnv}.

\noindent The \ac{dsc} quantifies the spatial overlap between the predicted segmentation mask \(P\) and the ground truth lesion mask \(G\). It is computed as:

\[
\text{\ac{dsc}} = \frac{2 |G \cap P|}{|G| + |P|}
\]

\noindent where \(|\cdot|\) denotes the cardinality, i.e., the number of voxels within the respective mask.

\noindent The \ac{fpv} quantifies the volumetric burden of predicted lesions that do not overlap with any ground truth lesion. To compute \ac{fpv}, the predicted mask \(P\) is decomposed into its connected components \(\{P_l\}_{l=1}^{L_P}\) with connectivity=18, each representing a distinct predicted lesion. For each predicted lesion \(P_l\), it is checked whether it overlaps with the ground truth mask \(G\). The volumes of all predicted lesions with no overlap are summed and scaled by the voxel volume \(v\), resulting in:

\[
\text{\ac{fpv}} = v \sum_{l=1}^{L_P} |P_l| \cdot \mathbf{1}(|P_l \cap G| = 0)
\]

\noindent where \(\mathbf{1}(x = 0)\) denotes the indicator function, defined as:
\[
\mathbf{1}(x = 0) =
\begin{cases}
1, & \text{if } x = 0 \\
0, & \text{otherwise}
\end{cases}
\]

\noindent The \ac{fnv} is defined analogously, quantifying the volumetric burden of ground truth lesions that do not overlap with any predicted lesion. The ground truth mask \(G\) is decomposed into its connected components \(\{G_i\}_{i=1}^{L_G}\), representing individual lesions. For each ground truth lesion \(G_i\), it is checked whether it overlaps with the predicted mask \(P\). The volumes of all missed lesions are summed and scaled by the voxel volume \(v\), yielding:

\[
\text{\ac{fnv}} = v \sum_{i=1}^{L_G} |G_i| \cdot \mathbf{1}(|G_i \cap P| = 0)
\]

\noindent For all metrics, the voxel volume \(v\) is assumed to be identical for both the predicted and ground truth segmentations.

\noindent If a case contains no tracer-avid lesions (i.e., if the ground truth mask is empty, \(G = \emptyset\)), the \ac{dsc} and \ac{fnv} cannot be computed. In such cases, only the False Positive Volume is evaluated.

\paragraph{Critique and justification}
The \ac{dsc} is computed exclusively on lesion-positive samples to avoid artificial metric inflation: in lesion-free cases, even a minor false positive would shift the \ac{dsc} from 1 to 0, disproportionately influencing the aggregated score. This is particularly relevant given that the "healthy" cohort largely comprises post-therapy patients in whom small spurious predictions are not uncommon. Because restricting the \ac{dsc} to positive cases removes false positives from the evaluation, we introduce the \ac{fpv} to explicitly quantify them. Symmetrically, we include the \ac{fnv} to capture missed lesions. Together, \ac{fpv} and \ac{fnv} act as volume-weighted detection metrics: by coupling detection with lesion volume, they ensure that small missed or spurious lesions carry proportionally less influence during patient-level aggregation, while large false negatives and false positives are penalized more heavily. The detection threshold is set to a single voxel. While permissive, this criterion reflects the clinical reality that lesion boundaries are inherently ambiguous and that a large predicted region encompassing a 
cluster of smaller annotations constitutes a valid detection. We will conduct additional post-challenge analyses to assess the sensitivity of the results to stricter detection thresholds.

\subsubsection{Ranking}
Participants were ranked according to the following scheme: We divided the test dataset into four subsets based on center and tracer (i.e., \uktfdg, \uktpsma, \lmufdg, \lmupsma) and calculated the average metrics for \ac{dsc}, \ac{fpv}, and \ac{fnv} within each subset. Then, we ranked the subset averages across all algorithms. For each metric, we computed an intermediate average rank by averaging the ranks of the subsets. Finally, we generated the overall rank by combining the three metric ranks using the following weighting factors: \ac{dsc} (0.5), \ac{fpv} (0.25), and \ac{fnv} (0.25). After that, teams were positioned by using the rank of the higher-performing algorithm. This procedure is done first for all teams participating in \ac{ac2}. All teams that rank higher than our data-centric baseline are eligible. For \ac{ac1}, all algorithms are ranked, including the algorithms submitted to \ac{ac2}. Baseline algorithms were also ranked, but excluded from awards. 

\subsubsection{Baseline algorithms}
Two baseline algorithms were provided as Docker containers to familiarize participants with the submission format and as a fixed algorithm for \ac{ac2}. The first was based on the nnU-Net framework \citep{isensee_nnu-net_2021}, which was applied out of the box to the training dataset. The second algorithm constituted a replica of the configured nnU-Net implemented in MONAI. We opted for this approach to gain control over which configuration parameters can be adapted and which cannot. The data-centric approach also includes a dynamic test time augmentation post-processing strategy as an exemplary modification for participants. Due to the five-minute time limit per sample, this approach calculates the number of test augmentations based on the prediction time of one sample, incorporating a safety threshold. 
In addition, after the challenge, we trained two nnU-Net baseline models for each tracer. These are used to reference single model performances and, when combined, serve as a virtual example of a perfect multi-stage routing approach.

\subsubsection{Post challenge analyses}
The post-challenge analysis is two-fold. The first part was conducted to assess the reliability and validity of the challenge results. We performed experiments on ranking stability and the influence of different ranking methods, mostly following \citet{wiesenfarth_methods_2021}. We present alternative metrics, ensembling results and a small reader analysis. 
The second part is insight-driven and focused on the main clinical tasks under a compositional setting: Volume estimation and lesion detection. For that, we split the analysis into patient-level and lesion-level components. 
The analysis is mostly descriptive however to account for the hierarchical data structure, we also employ linear mixed-effects models. 

\begin{figure*}[!ht]
\centering
\includegraphics[width=\textwidth]{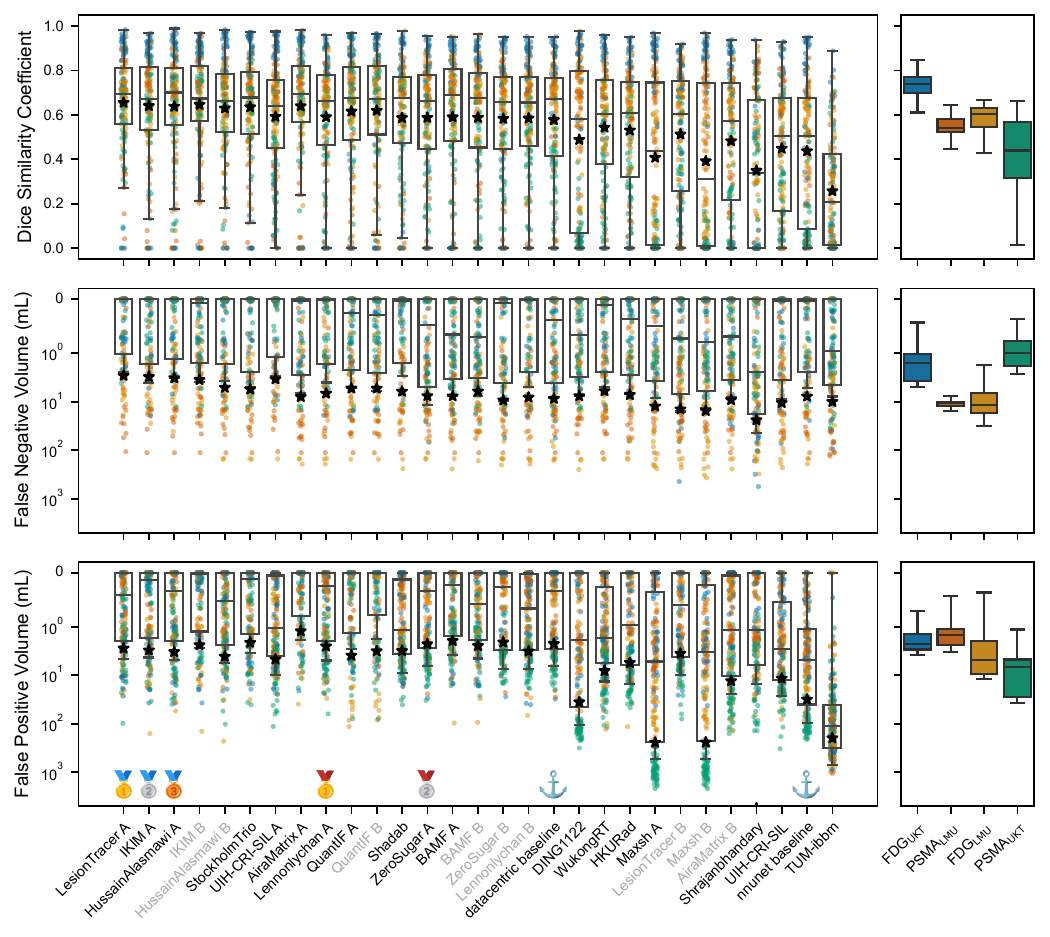}
\phantomsection
\caption{Performance of all 29 submitted algorithms across the four test conditions, ordered left to right by final leaderboard position. Each column shows the results for one algorithm as as a boxplot; rows show the \ac{dsc} (top), \ac{fnv} (middle), and \ac{fpv} (bottom). \ac{fpv} and \ac{fnv} are displayed on a symmetric logarithmic scale. Individual test cases are shown as points, colored by domain. Stars indicate the mean across all data points. Blue medals correspond to the winning teams in \ac{ac1}, red ones to those in \ac{ac2} and anchors indicate the baselines. The right side shows the distribution of team averages per-domain.}
\label{fig_overview_results}
\end{figure*}

\section{Results}
\subsection{Challenge Participation}
As of November 2024, a total of 563 individuals had registered for the \ap challenge. The participants were predominantly from Asia (53\%, with 29\% from China), followed by Europe (28\%, primarily Germany at 8\%) and North America (13\%, mainly the United States at 10\%). Oceania (2\%), South America (2\%) and Africa (\textless 1\%) were underrepresented.  
During the preliminary phase, 24 teams participated, submitting a total of 218 algorithms. In the final phase, 17 teams submitted 27 algorithms, with the highest-performing submission from each team being used to determine the final leaderboard position. Four teams submitted a total of six algorithms to \ac{ac2}. Two teams submitted exclusively to this category, and two teams submitted one model each to \ac{ac1} and \ac{ac2}.

\subsection{Submitted algorithms}
\label{sec:algo-overview}
Almost all participating teams in \ac{ac1} utilized the nnU-Net framework \citep{isensee_nnu-net_2021} (12/15). All teams handled the multi-modality by concatenating CT and PET volumes. Only one submitted algorithm used a 2D Unet for segmentation (AiraMatrix), while all others used 3D models. We observed two different architectural flows for handling multitracer data. 11 teams submitted a single model for handling both tracers, and four teams (IKIM, UIH-CRI-SIL, QuantIF, Maxsh) proposed a two-stage approach where a routing network predicts the tracer and routes the sample to an expert model. One team (HKURad) used a multistage approach, combining a coarse segmentation network to identify relevant regions of interest with fine segmentation. Six teams used organ masks produced by the TotalSegmentator \citep{wasserthal_totalsegmentator_2023}. One team (AiraMatrix) used organ masks solely to crop the body to a relevant FOV. The remaining teams (LesionTracer, BAMF, IKIM, UIH-CRI-SIL, QuantIF) went a step further and incorporated selected organ labels directly into their training. All of them shared a core set of high-uptake organs (liver, kidneys, bladder, spleen, lungs, brain, heart, prostate, and stomach), but differed in the additional structures they included. BAMF added rarer anatomies such as the adrenal glands, thyroid, and skull. Only LesionTracer included salivary glands. IKIM varied its organ subset depending on the tracer. UIH-CRI-SIL included the femurs. QuantIF added the aorta and grouped structures into coarser categories. Notably, no two teams used exactly the same organ label set.

Most teams used standard nnU-Net 5-fold ensembling. An exception was AiraMatrix, which used the STAPLE algorithm \citep{warfield_simultaneous_2004} for combining fold predictions. Several teams (LesionTracer, AiraMatrix, QuantIF, WukongRT) implemented dynamic test-time augmentation, adapting the number of mirroring axes based on available inference time within the 5-minute challenge constraint.
For PET normalization, most algorithms used a global normalization scheme based on the nnU-Net preprocessing (LesionTracer, HussainAlasmawi A, StockholmTrio, AiraMatrix, DING1122, HKURad, Maxsh). Several teams used per-image z-score normalization for PET (IKIM, HussainAlasmawi B, UIH-CRI-SIL, QuantIF, BAMF, WukongRT, TUM-ibbm), min-max normalization (Shadab), or experimented with novel normalization techniques (WukongRT, Shrajanbhandary).

Backbone sizes varied across teams. Most used the nnU-Net default ResUNet (Shadab, DING1122, Shrajanbhandary, TUM-ibbm). Others went with newer nnU-Net backbone presets \citep{linguraru_nnu-net_2024}. IKIM and WukongRT chose a medium Residual Encoder (ResEnc M) variant, while LesionTracer, HussainAlasmawi, and StockholmTrio adopted the larger ResEnc L. BAMF and AiraMatrix used the largest available backbone (ResEnc XL). Patch sizes ranged accordingly: teams with smaller backbones typically trained on patches around $128^3$ voxels (e.g. IKIM: $128\times112\times160$; UIH-CRI-SIL: $128^3$), whereas larger backbones were paired with patches near or above $192^3$ (e.g., LesionTracer and HussainAlasmawi: $192^3$; BAMF: $256\times256\times192$; AiraMatrix: $224\times192\times224$).

In \ac{ac2} methodological modifications were limited to data handling. Lennonlychan enlarged the dataset by using a diffusion model to generate tumor samples. ZeroSugar filtered samples, LesionTracer B introduced a new misalignment augmentation, and UIH-CRI-SIL B experimented with Gaussian sharpening and normalization with clipping. 

We summarize the methods of the top three methods of \ac{ac1} and the top two methods of \ac{ac2} in detail in \ref{sec:topalgo}.

\begin{figure*}[!ht]
\centering
\includegraphics[width=\textwidth]{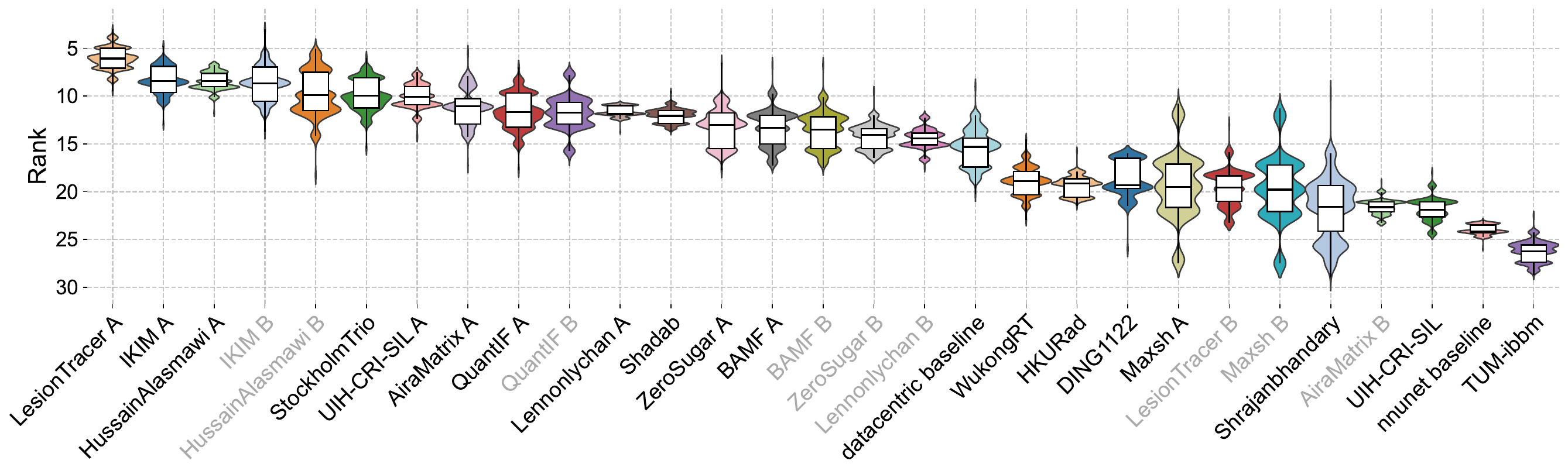}
\caption{Bootstrap ranking stability analysis (n=2,000) of all submitted algorithms, shown as violin plots. Lower rank indicates better performance. A clear performance gap is visible after the data-centric baseline, separating algorithms into two tiers. Among the top-performing group, LesionTracer A achieves the most consistent top ranking, while several mid-tier algorithms show overlapping rank distributions, indicating comparable performance. 
}\label{fig_ranking_stability}
\end{figure*}

\subsection{Performance of submitted algorithms}
Figure~\ref{fig_overview_results} summarizes the performance of all submitted algorithms. In addition, details on individual performance can be found in Table~\ref{tab:raw_results} and Figure \ref{fig:individual_metrics}.  LesionTracer~A achieved the highest overall position (1st),
followed by IKIM~A and HussainAlasmawi~A. The top three algorithms demonstrated balanced \ac{dsc} performance across tracers and centers, with low \ac{fnv} ranks and slightly higher \ac{fpv} ranks. AiraMatrix~A (6th) achieved the best \ac{fpv} rank, driven by comparably low false positives on the \ac{psma} datasets, though at the cost of higher \ac{fnv}.
 The data-centric baseline serves as a natural boundary between a large group of well-performing algorithms and those with less stable predictions. Among the well-performing group, median \ac{dsc} is remarkably similar across teams (range 0.64-0.70). Clearly visible is a zero inflation in \ac{dsc} and large tails for \ac{fpv} and \ac{fnv}. 
The three metrics reveal complementary aspects of generalization. The best \ac{dsc} was obtained on \uktfdg and the worst on \uktpsma, with \lmupsma and \lmufdg in between. The most severe false negatives (10--100\,mL) occurred predominantly on the LMU datasets.
In contrast, on \ac{fpv}, most algorithms produced fewer false positives on in-domain data than on their composites. 
In \ac{ac2}, only two teams achieved slightly better performance than the data-centric baseline: Lennonlychan (1st, 7th overall) and ZeroSugar (2nd, 10th overall). Our data-centric baseline outperformed the standard nnU-Net version.

\subsubsection{Multi-stage models}
\label{sec:multistage}
As a small-scale ablation, we evaluated the quality of the tracer-routing models for three of the four teams. Team IKIM and QuantIF were inferred using the provided code, and for team UIH-CRI-SIL, we used the challenge predictions' logs.  
IKIM's classifier misclassified only five of 200 samples (Accuracy 97.5). The UIH-CRI-SIL tracer model performs even better, misclassifying only one sample. Team QuantIF reached an Accuracy of 0.93 with 14 errors. For all teams, all misclassifications were on the \lmufdg dataset, which was not part of the training set. 

\subsubsection{Ensembling}
\label{sec:ensembling}
We explore the potential of combining the top five-ranked team algorithms via majority-vote ensembling (LesionTracer A, IKIM A, HussainAlasmawi A, StockholmTrio, UIH-CRI-SIL A). The ensemble achieved competitive performance as shown in Table \ref{tab:raw_results}.
While individual algorithms still outperform the ensemble on specific subgroups, the ensemble averages out weaknesses across the remaining datasets. In particular, \ac{fpv} is drastically reduced compared to the individual algorithms, while \ac{dsc} and \ac{fnv} remain competitive. If the ensemble had been ranked alongside the participating algorithms, it would have secured first position (weighted rank 4.56), achieving the best \ac{dsc} and \ac{fnv} ranks (3.50 and 3.00, respectively) and the third-best \ac{fpv} rank (8.25) after AiraMatrix and BAMF. 

As an additional experiment, we combined the two additional baseline models, each trained exclusively on a single tracer, simulating a perfect routing that always selects the correct specialist. Even under this optimistic assumption, the combined model would only rank 10th, performing worse than three other expert models (IKIM, UIH-CRI-SIL, QuantIF).

\subsection{Challenge reliability and validity}
\subsubsection{Ranking stability with respect to sampling variability}
The robustness of the proposed ranking with respect to the test data was assessed using a bootstrap analysis (n=2000) following \cite{wiesenfarth_methods_2021}. The results in Figure \ref{fig_ranking_stability} demonstrate the distribution of the rankings. A clear drop is visible after the data-centric baseline, separating the algorithms into two groups. Further, the first algorithm, LesionTracer A, consistently outranks the others, except for some overlap with IKIM A and B. IKIM, on the other hand, shows greater variation in rankings. Many of the following algorithms have overlapping intervals or even similar median ranks, indicating performances on par (HussainAlasmawi B, StockholmTrio, and UIH-CRI-SIL A).

\subsubsection{Ranking stability with respect to different ranking methods}
\label{sec:altrank}
\begin{figure}[t]
\centering
\includegraphics[width=0.5\textwidth]{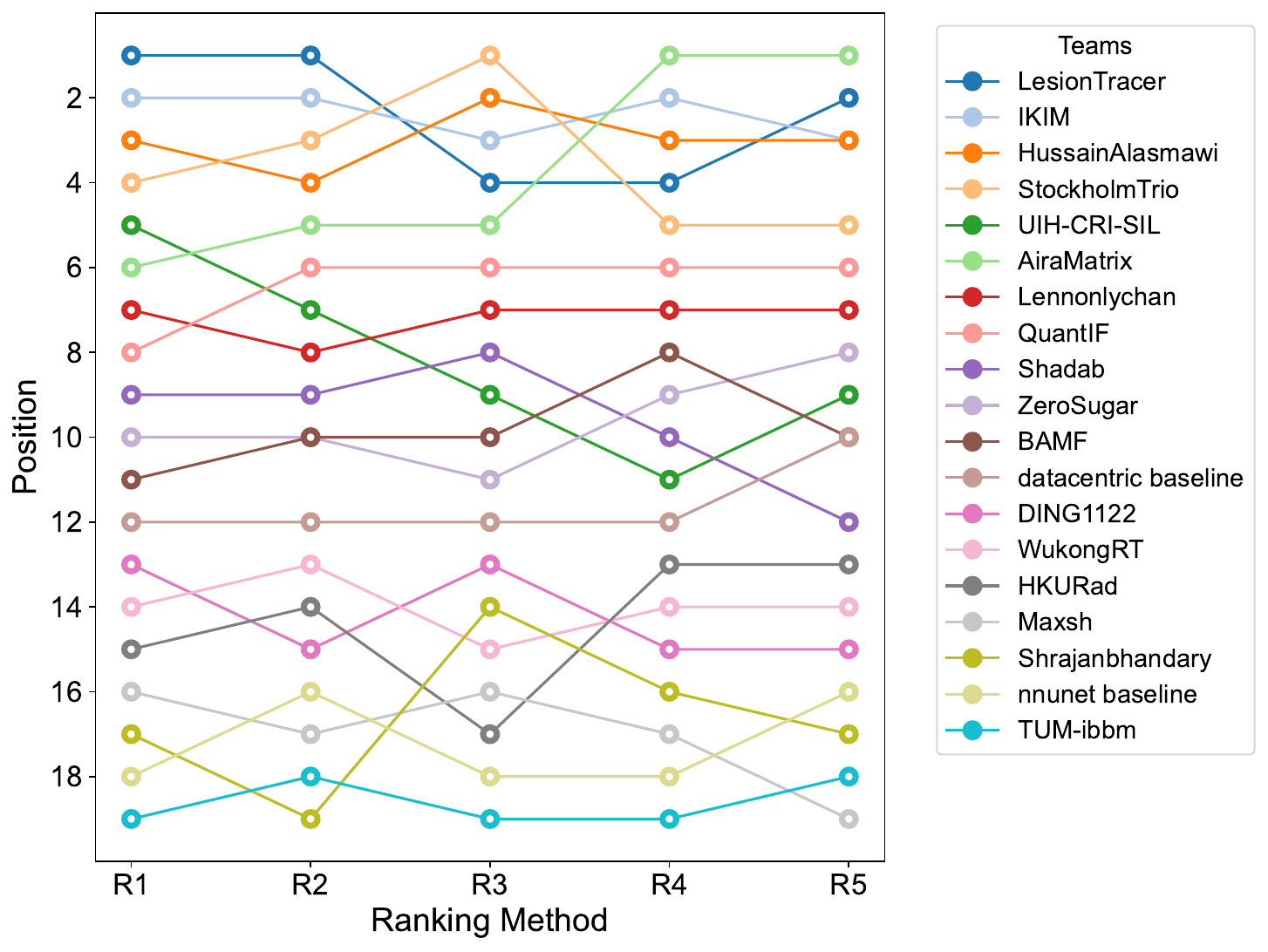}
\caption{Ranking stability across five ranking methods: official ranking (R1), aggregate-then-rank (R2), median-per-subgroup (R3), rank-then-aggregate (R4), and test-then-rank (R5). Each line represents a team, with position on the y-axis (lower is better). The best algorithm is used for determining the position. Top- and bottom-tier teams remain stable, while mid-field positions (4–12) show the greatest variability.}
\label{fig:alternative_rankings}
\end{figure}

To assess the robustness of the challenge ranking, we evaluated five ranking methods. The official ranking (R1) computes the mean metric value per subgroup, ranks teams within each subgroup, and averages ranks across the four subgroups. Aggregate-then-rank (R2) averages the subgroup means into a single score and ranks on that aggregate. Median-per-subgroup (R3) replaces the mean within each subgroup with the median before ranking, thereby providing robustness to outliers. Rank-then-aggregate (R4) first ranks algorithms per test case, then averages case-level ranks. Test-then-rank (R5) performs pairwise Wilcoxon signed-rank tests between all algorithm pairs within each subgroup, applies Holm correction for multiple comparisons, and ranks teams by the number of opponents they significantly outperform (p \textless~0.05).
Figure \ref{fig:alternative_rankings} shows the resulting ranking stability plot. The top-ranked teams (LesionTracer, IKIM, HussainAlasmawi) maintain positions 1-4 across all five methods, demonstrating that their superiority is robust to the choice of ranking scheme. Similarly, the bottom tier (positions 15–19) remains stable. The greatest variability occurs in the mid-field (positions 4–12), where teams such as UIH-CRI-SIL, AiraMatrix, and StockholmTrio shift by up to 5 positions depending on the method. 
The data-centric baseline again separates the field into two groups, with no algorithm from the lower tier crossing above it under any ranking method.

\subsubsection{Additional performance metrics}
\label{sec:addmetrics}
We report several additional metrics in Table~\ref{tab:alternative_metrics}, following the approach of previous autoPET editions. For simplicity, we use a single weighted average across the four test conditions.
In addition to the challenge metrics, we include two complementary voxel-level measures. The normalized surface distance (NSD)~\citep{nikolov_clinically_2021} quantifies boundary agreement at a tolerance of one voxel, capturing contour accuracy independently of volume. Volumetric similarity (VS)~\citep{taha_metrics_2015} measures the relative difference in predicted and reference volumes without penalizing spatial displacement, providing a volume-focused counterpart to \ac{dsc}.

At the lesion level, we report two instance-aware metrics. Connected-component \ac{dsc} (CC-DSC)~\citep{jaus_every_2025} evaluates \ac{dsc} on a per-lesion basis by assigning predictions to their nearest ground-truth component via a proximity-based Voronoi partition, averaging across all components irrespective of their size. Panoptic quality (PQ)~\citep{kirillov_panoptic_2019} is reported at an IoU threshold of $\tau = 0.1$, alongside its composites recognition quality (RQ) and segmentation quality. RQ is equivalent to the F1 score under the given matching threshold, reflecting detection performance, while SQ reflects the mean IoU of matched pairs (multi-assignment is not penalized, and for SQ, the lesion with the highest overlap is used).

All lesion-level and voxel-level metrics are computed exclusively on lesion-positive cases. To illustrate the effect of resolving empty predictions, which are conventionally assigned a \ac{dsc} of either zero or one, we additionally report \ac{dsc} computed over all samples. As a global detection metric, we report the F1 score aggregated across all samples. 

\begin{figure}[t]
\centering
\includegraphics[width=0.5\textwidth]{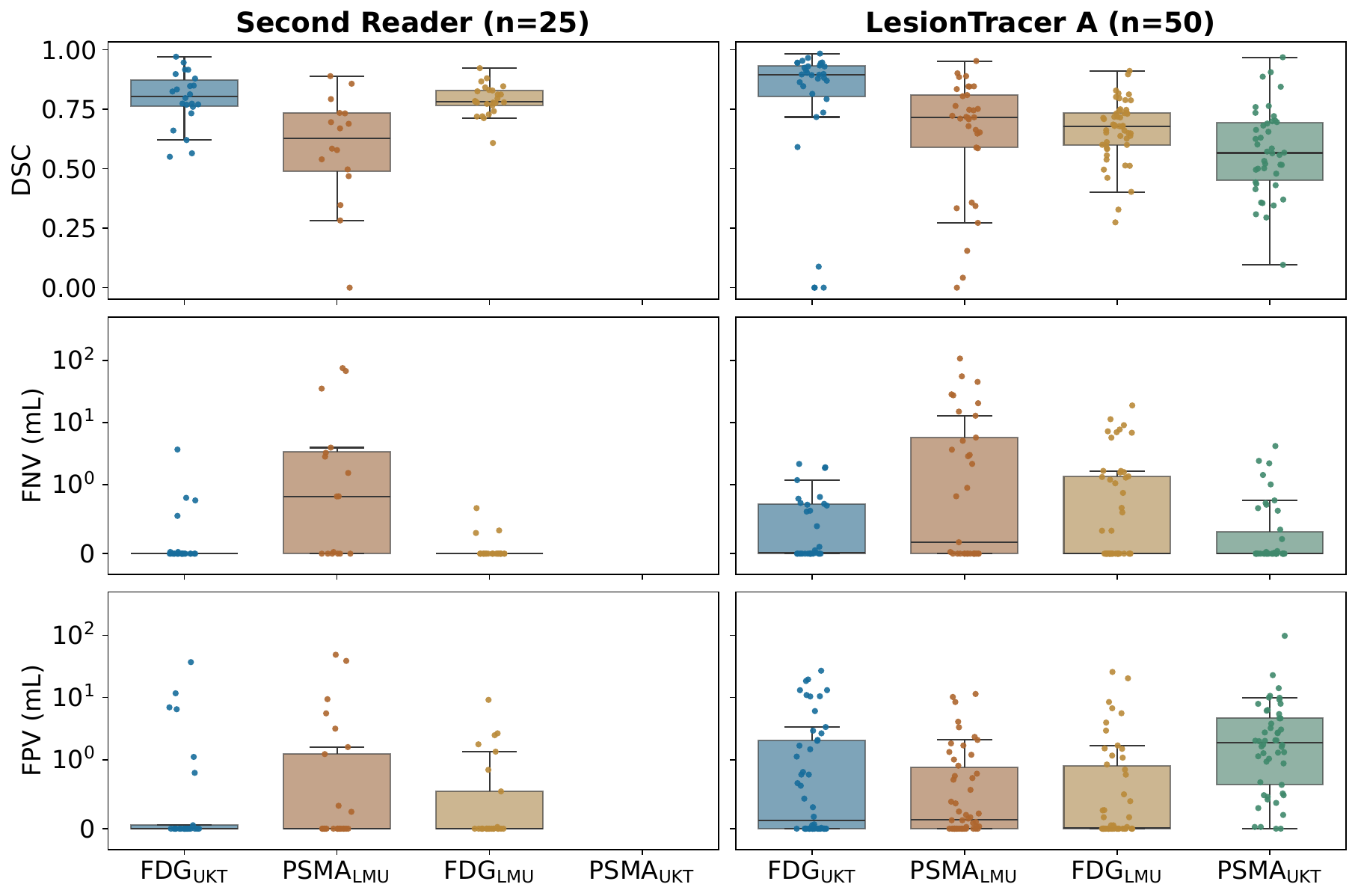}
\caption{Comparison of the top-ranked algorithm (LesionTracer A, n=50) with second-reader agreement (n=25) in terms of \ac{dsc} (top), false-negative volume (middle), and false-positive volume (bottom, both in mL, symlog scale) across the four test conditions. Reader subsets were drawn from the same distribution but are not identical to the algorithm test sets, and reading protocols varied across conditions (see main text for details). No second-reader data were available for \uktpsma.
}
\label{fig:reader_variability}
\end{figure}

\begin{figure*}[!ht]
\centering
\includegraphics[width=\textwidth]{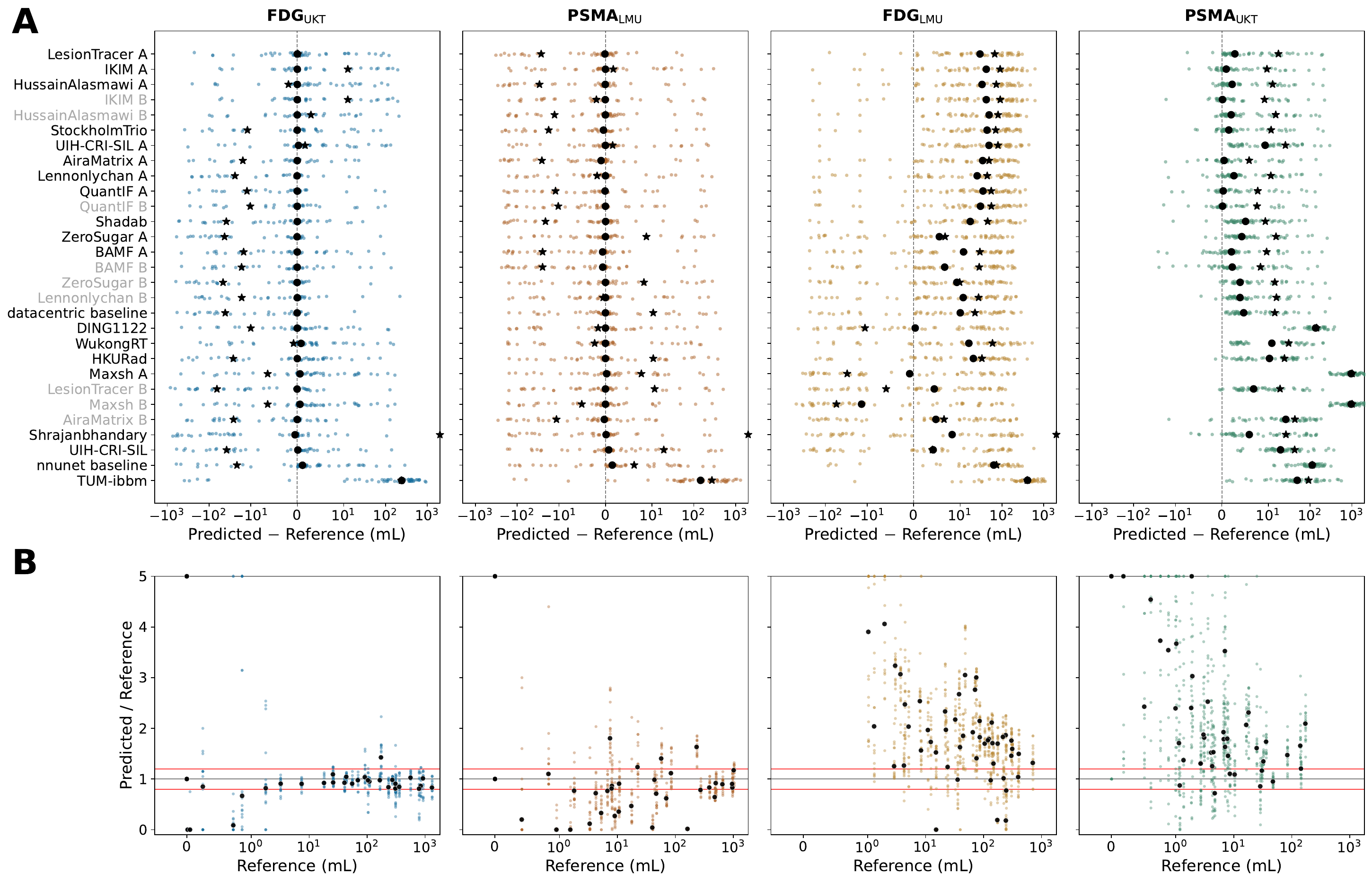}
\caption{Volume analysis across all datasets. \textbf{(A)}~Per-algorithm absolute volume difference (predicted $-$ reference, mL) displayed on a symmetric log scale. Large black circles indicate the median, stars denote the mean, and individual colored dots represent per-case differences. Algorithms oversegment in the composite datasets.
\textbf{(B)}~Relative volume agreement for the top-18 algorithms, shown as $(\text{pred} + \epsilon) / (\text{ref} + \epsilon)$ plotted against reference volume, where $\epsilon=0.012$. Black circles indicate median ratios per case across algorithms; colored dots show individual algorithm predictions. Red curves denote relative $\pm 20\%$ boundaries.}
\label{fig_volume}
\end{figure*}

\subsubsection{Reader agreements}
\label{sec:reader}
We compared the top-performing algorithm (LesionTracer A) against available reader segmentations, which should be interpreted as approximate reference points rather than rigorous inter-reader benchmarks, given the heterogeneous conditions under which they were obtained. For \uktfdg and \lmufdg, second readings were drawn from the first autoPET challenge \citep{gatidis_results_2024}, where the same experienced reader (S.G., 10+ years in hybrid imaging) re-annotated 25 cases after a washout period. For \lmupsma, a junior reader (G.A.) independently annotated 25 randomly selected test-set cases; no second-reader data were available for \uktpsma. Figure~\ref{fig:reader_variability} shows agreement in terms of \ac{dsc}, \ac{fnv}, and \ac{fpv}. For \uktfdg and \lmupsma, LesionTracer A achieved performance comparable to or exceeding the second reader across all three metrics. In contrast, for \lmufdg, the best algorithm remained below the second-reader agreement.
We were surprised by the low inter-rater agreement in \lmupsma and verified most disagreeing cases with an experienced radiologist (M.H.F., 6 years of experience). These samples contained only lesions where annotation is difficult and inherently ambiguous, driving the metrics. For example, one error involved a small local recurrence that could also be interpreted as residual urine, yet could arguably be counted as a lesion since the bladder was empty.

\subsection{Patient-level analysis}
\subsubsection{Factors driving segmentation performance}
\label{sec:mixedmodel}
Figure~\ref{fig:sample_dist} shows \ac{dsc}, \ac{fnv}, and \ac{fpv} distributions per patient across the top-18 algorithms, sorted by average \ac{dsc}. Patient-level heterogeneity is substantial, with median \ac{dsc} ranging from zero to above 0.9. A small cluster on the left shows near-zero \ac{dsc} across almost all teams. Visual inspection reveals that all of these cases have only one small lesion. The right side is dominated by \uktfdg samples with high median scores and narrow interquartile ranges, while \lmufdg patients populate the mid-range, and \lmupsma is scattered across the full spectrum. \uktpsma consistently shows larger variance and lower \ac{dsc}. There are also individual patients scattered across the range where algorithms substantially disagree.

Across top-18 algorithms, performance varied far more between patients than between teams, a pattern we quantified using a linear mixed-effects model (\ref{sec:apmixed}, Model 1). 
The variance partition reveals a clear hierarchy: patient heterogeneity accounts for 61\% of the unexplained variance in \ac{dsc} ($\sigma_{\text{patient}} = 0.171$), the residual for 38\% ($\sigma_{\text{residual}} = 0.134$), and mean algorithmic differences for only 1.3\% ($\sigma_{\text{team}} = 0.025$), even after accounting for tracer, center, their interaction and tumor volume.
In concrete terms, the expected \ac{dsc} difference between two randomly selected patients, averaged across algorithms, is roughly seven times as large as the difference between two randomly selected algorithms, averaged across patients.
Among the fixed effects, reference lesion volume was the strongest predictor: each doubling was associated with a $+0.039$ increase in \ac{dsc} $[0.030,\, 0.048]$. After volume adjustment, a center effect emerged (UKT $+0.14$ $[0.063,\, 0.217]$), while neither tracer nor the tracer$\times$center interaction reached significance. 

\subsubsection{Ablation \ac{psma} ligands}
\label{sec:ligands}
We fitted a small linear model to determine if there is a bias towards different tracer ligands (Appendix A, Model 2).
Within the \ac{psma}-LMU subset, no performance difference was observed between \textsuperscript{68}Ga-\ac{psma}-11 (n=16) and \textsuperscript{18}F-\ac{psma}-1007 (n=17)  ($\beta = -0.03$, 95\% CI [−0.21, 0.16]), indicating that algorithms generalize across \ac{psma} radioligands despite their known differences in biodistribution.

\subsubsection{Patient-level classification}
\label{sec:classification}
To verify whether the algorithms can flag patients with pathological uptake, we report the number of true-positive and true-negative cases in Table~\ref{tab:alternative_metrics}. Across all teams, true-positive counts were consistently high (range: 136--152 of 156 positive cases), whereas true-negative counts varied substantially (range: 0--35 of 44 negative cases). Performance differed between algorithms. LesionTracer A was the most sensitive (sensitivity = 0.97, specificity = 0.27, accuracy = 0.82) while AiraMatrix A the most specific (sensitivity = 0.92, specificity = 0.8, accuracy = 0.90).

\subsubsection{Volume estimation}
\label{sec:volume}
Accurate TMTV estimation is clinically important for treatment stratification.
Figure~\ref{fig_volume}\,(A) shows the volumetric differences per team. It is visible that in the two in-domain datasets, the median of almost all algorithms is close to zero. There are, however, substantial outliers dragging the means differently across teams. For out-of-domain datasets, the median and mean shift towards oversegmentation.

This becomes very apparent in Figure~\ref{fig_volume}\,(B) where the relative change of the top 18 algorithms is plotted. The red lines indicate a $\pm\,20\%$ boundary. For the in-domain data, many algorithms predict volumes within this range. \lmupsma, however, shows notable undersegmentations in the range of 1mL to 100mL. The out-of-domain datasets are systematically shifted towards oversegmentation (with roughly 1.7 times oversegmentation) even for cases with large reference volume. In addition, the variance within algorithms is much larger. Many cases without any reference volume also have an associated volume; however, the magnitude is relatively low.

\subsection{Lesion-level analysis}
\subsubsection{Lesion detection}
\label{sec:detection}
\begin{figure}[!t]

\centering
\includegraphics[width=\columnwidth]{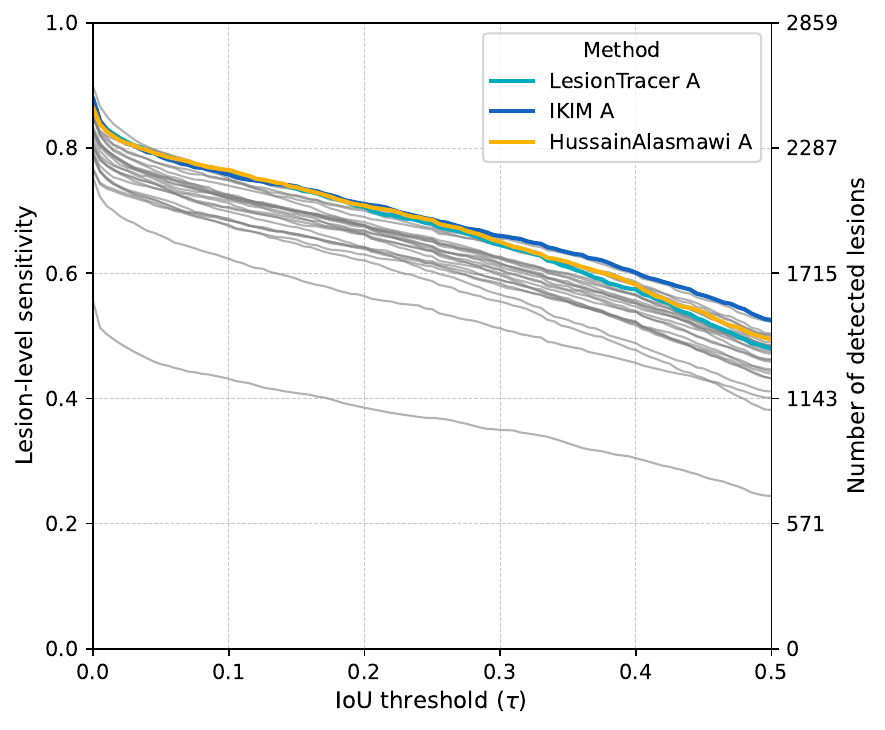}
\caption{Lesion detection sensitivity as a function of the IoU threshold $\tau$. The left end of the abscissa corresponds to the one-voxel criterion; the right end to $\tau = 0.5$, analogous to the recognition in panoptic quality~\citep{kirillov_panoptic_2019}, which enforces one-to-one matching. The figure is based on an assignment strategy that does not penalize multi-assignment. Top 3 teams are highlighted.}\label{fig_lesion_dect}
\end{figure}

The second clinically relevant question is lesion detection: are all true lesions found? While the challenge metrics use a simple one-voxel matching criterion for \ac{fnv} and \ac{fpv}, detection performance in practice depends on the overlap metric, threshold, and matching strategy (i.e., how to handle multi-assignment). Figure~\ref{fig_lesion_dect} shows lesion detection sensitivity across all teams as a function of the IoU threshold $\tau$.

Overall median sensitivity approximately halves from 0.83 at the one-voxel criterion to 0.48 at $\tau = 0.5$, and decreases monotonically in between. The sharp rise near the one-voxel threshold suggests that many detections rely on only marginal overlap. Better-performing algorithms tend to maintain higher sensitivity across the entire range, though some intersections between algorithms are visible. 

Grouping by center and tracer reveals notable differences (Figure \ref{fig:detection_deciles_per_ds}). Three of the four dataset conditions start above 0.84 at the one-voxel criterion, whereas \lmufdg reaches only 0.74 and declines more steeply across the threshold range. The rapid decay from the one-voxel threshold is also more pronounced for \lmufdg\ than for \lmupsma, and more algorithm intersections become visible in the per-dataset view.

\begin{figure*}[!ht]
\centering
\includegraphics[width=\textwidth]{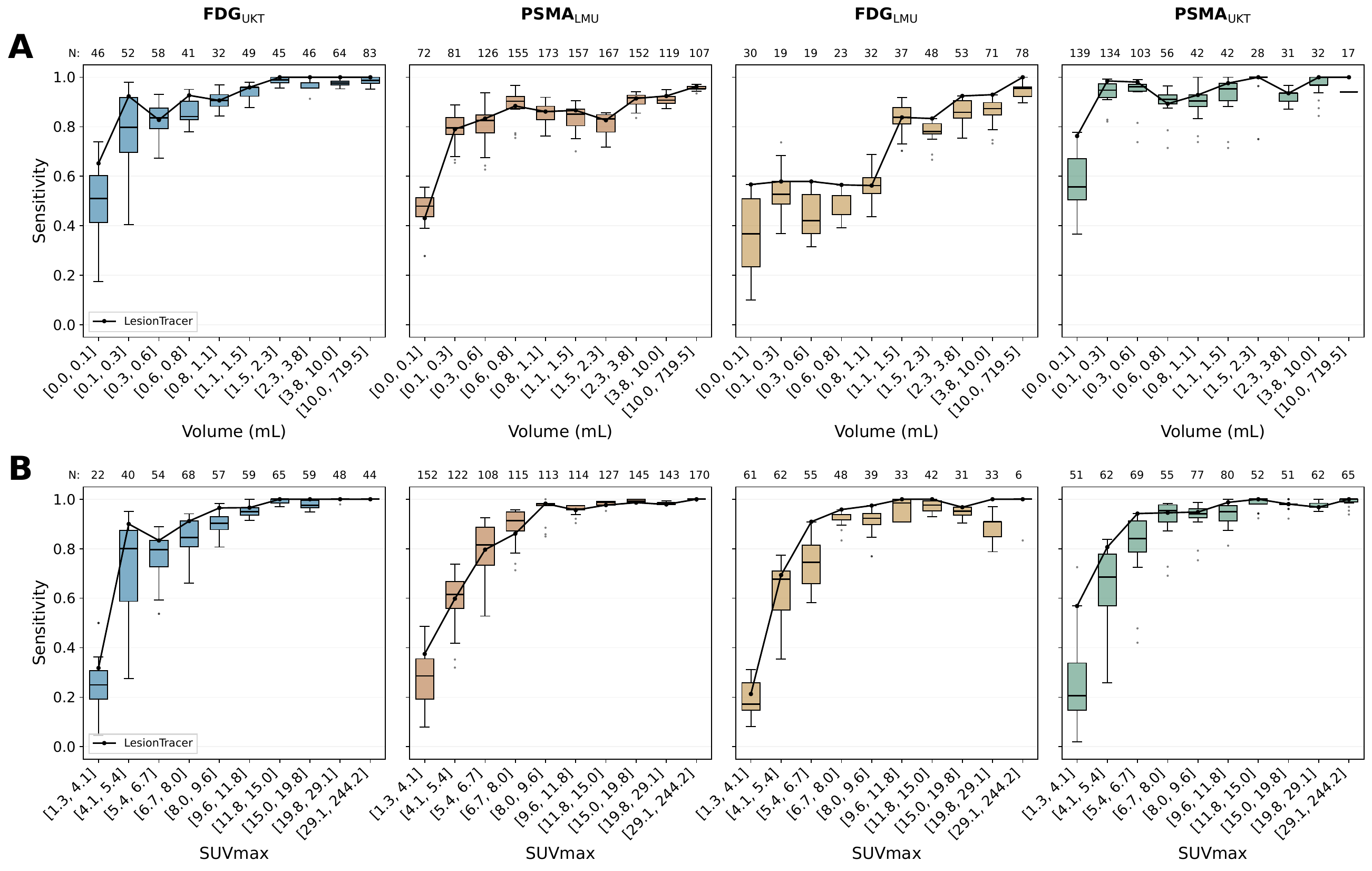}
\caption{Lesion detection sensitivity stratified by volume deciles (A) and SUV\textsubscript{max} deciles (B) across the four test conditions. Box plots summarize the distribution of per-algorithm sensitivity within each bin; the black line indicates the top-ranked team (LesionTracer).
Detection sensitivity increases with both lesion volume and tracer uptake across all conditions.
 }\label{fig:lesiondeciles}
\end{figure*}

\subsubsection{Detection errors}
\label{sec:lesionerr}
The simple detection/miss view hides structural errors: a single ground-truth lesion may be split into several predictions, or multiple ground-truth lesions merged into one. Following \citet{nascimento_performance_2006} and \citet{carass_evaluating_2020}, we decompose all prediction–reference associations into mutually exclusive categories: correct detections (CD, one-to-one match), false alarms (FA), detection failures (DF), merges (M, one prediction covers multiple references), splits (S, one reference covered by multiple predictions), and split-merges (SM, both conditions simultaneously).

Figure \ref{fig:nascimiento} shows these error types across the threshold range. At the one-voxel threshold, a substantial number of merge (median 146), split (median 92), and split-merge (median 24) associations exist, meaning that part of the high sensitivity in Figure \ref{fig_lesion_dect} stems from ambiguous matchings rather than clean detections. As $\tau$ increases, these cluster associations decay toward zero resolving into correct detections, detection failures and false alarms. Because of that correct detections slightly increase, reaching a median peak at $\tau \approx 0.08$ (range: $[0.03{-}0.14]$) across all teams. At roughly $\tau = 0.3$, the increase in FP and FN volumes seems to happen more drastically, while almost all cluster associations are gone. Visual inspection of merge and split clusters at the one-voxel criterion reveals ambiguity in the reference annotations. Lesions are often in close proximity, separated or connected by only a few voxels. Changing the connectivity criterion alone results in different reference instances ($cc_{6}$=3069, $cc_{18}$=2859, $cc_{26}$=2800).

\subsubsection{Factors influencing lesion detection}
\label{sec:lesionsuv}
It is frequently noted in the literature that lesion size and tracer uptake are primary drivers of detectability.
Figure \ref{fig:lesiondeciles} shows per-algorithm lesion detection sensitivity stratified by volume deciles (A) and SUV\textsubscript{max} deciles (B) across the four test conditions. Detection rates increase with both lesion size and uptake in all conditions. The smallest lesions ($< 0.1$\,mL) are detected at lower rates (roughly 40-60\%), while the largest lesions approach near-perfect detection. The dependence on SUV\textsubscript{max} is even steeper: lesions with SUV\textsubscript{max} below 4.1 experience a strong detection penalty (less than 40\%), rising to above 95\% for SUV\textsubscript{max} above 15. Across three conditions, the curves look similar, while \lmufdg shows slightly lower detectability for lesions between 0.1 and 1.5mL. These marginal views, however, obscure the joint dependence between the two factors. Larger lesions show greater uptake, and vice versa. Figure~\ref{fig:suv_uncert} makes this interaction explicit: SUV\textsubscript{max} appears to be the dominant driver of detection at the one-voxel criterion. Top-18 teams largely agree on most cases, the greatest inter-team variability is concentrated in lesions with an SUV\textsubscript{max} between 4 and 8.

\begin{figure*}[!ht]
\centering
\includegraphics[width=\textwidth]{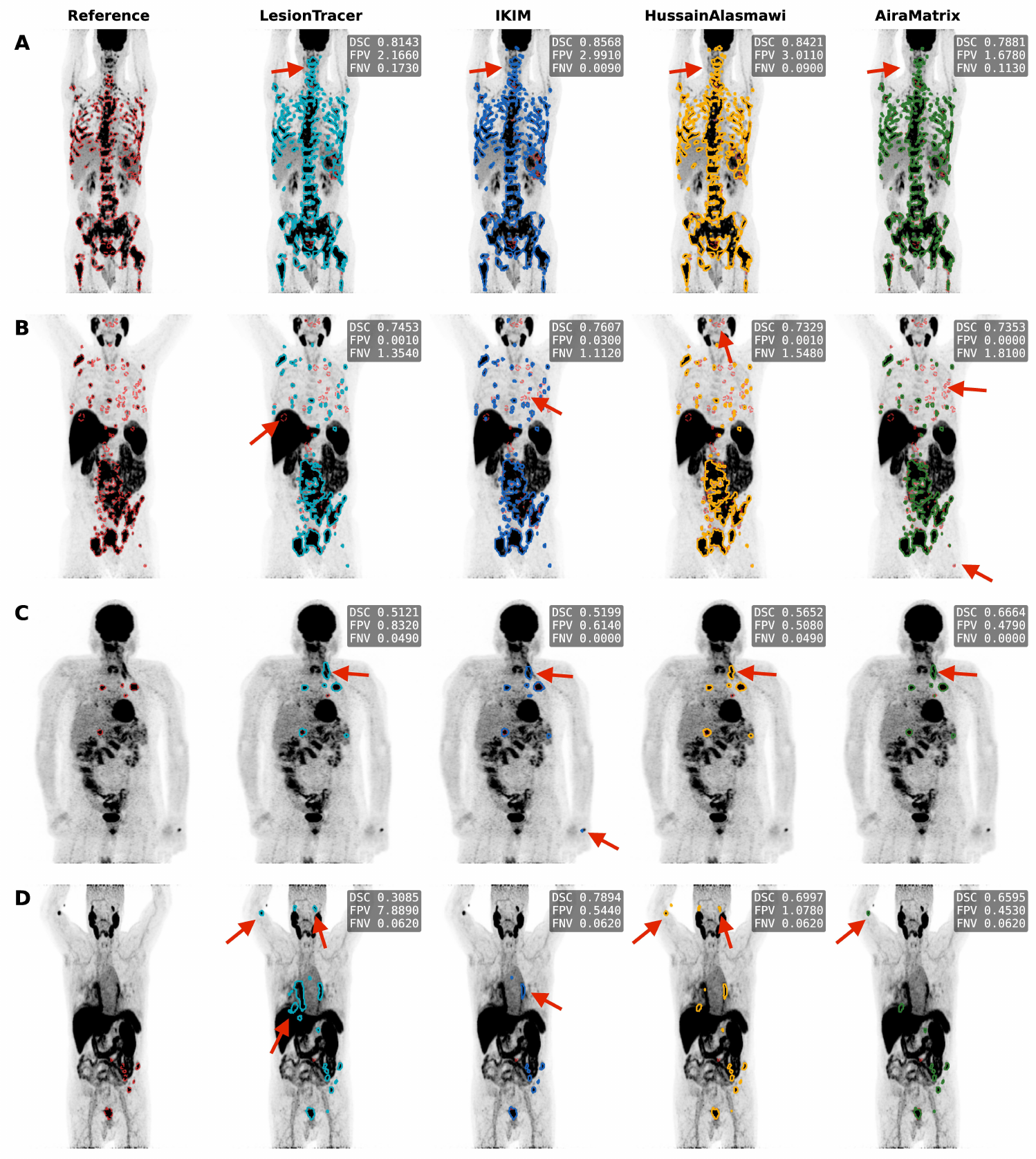}

\caption{Qualitative comparison of segmentation predictions from four algorithms (LesionTracer A, IKIM A, HussainAlasmawi A, AiraMatrix A) against the reference annotation on four representative failure cases, shown as coronal maximum intensity projections. Algorithm predictions are shown as colored contours overlaid on the reference (red). Metrics are reported for each algorithm. Failure modes are marked by arrows. The four panels illustrate distinct error categories: false positives from incomplete reference annotations in a \uktfdg case (A), systematic false negatives for low-expression lesions in a \lmupsma case (B), false positives from physiologic muscle uptake and injection-site infiltration in a \lmufdg case (C), and false positives from physiological tracer uptake and atelectasis in a \uktpsma case (D). Images show coronal SUVs with a window of [0, 7].
}
\label{fig_errors}
\end{figure*}

\begin{figure*}[!ht]
\centering
\includegraphics[width=0.9\textwidth]{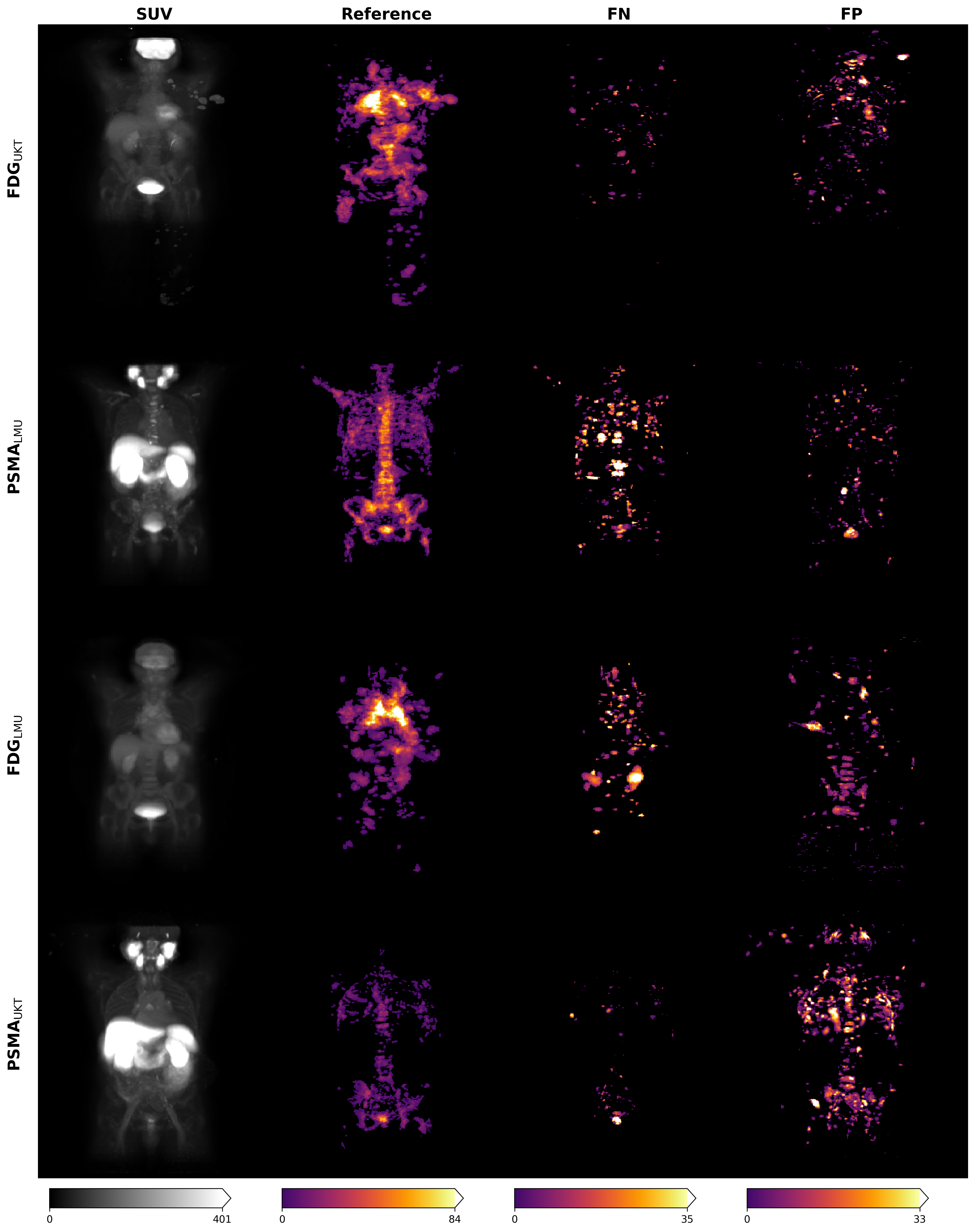}
\caption{To visualize where lesions, false positives, and false negatives are located across the test sets, we registered all cases to a common reference space. We first resampled every case to the median voxel spacing, then chose the largest-volume case as the reference and padded it to fit all anatomies. Elastic registration was performed using organ masks from TotalSegmentator on the CT images. The same transformations were applied to the SUV volumes, ground-truth annotations, and the summed false-positive and false-negative volumes of the top five teams (LesionTracer A, IKIM A, HussainAlasmawi A, StockholmTrio, UIH-CRI-SIL A). After summing all volumes, MIPs were produced. The result gives a qualitative sense of where ground-truth lesions are distributed and where the best-performing methods fail.}\label{fig_errors2}
\end{figure*}

\subsection{Error analysis}
Representative errors are visible in Figure \ref{fig_errors}. We plot the top 3 algorithms as well as the algorithm with the lowest \ac{fpv} (AiraMatrix A). Panel A shows an \uktfdg lung cancer case with extensive skeletal metastases. All algorithms exhibit elevated \ac{fpv} values (1.68–3.01), with predictions extending into skeletal regions where the reference annotation appears incomplete. Panel B depicts an \lmupsma case with progressive disease. Here, all algorithms show high \ac{fnv} (1.11–1.81) with near-zero \ac{fpv}. Several annotated lesions, particularly in the thoracic skeleton and cervical lymph nodes, show minimal tracer uptake. These low-expression lesions are systematically missed by all algorithms. Panel C shows an \lmufdg bronchial carcinoma case in which all algorithms produce elevated false-positive predictions (\ac{fpv} 0.48–0.83). The false positives are predominantly produced by physiologic muscle activity. Additionally, one algorithm predicts a false positive near the hand, consistent with tracer infiltration at the injection site. Panel D presents an \uktpsma case where false positives are observed in the lacrimal glands, a site of known physiological \ac{psma} expression, and in the region superior to the liver, consistent with dependent atelectasis. Three algorithms also produce additional false positives attributable to tracer infiltration at the injection site.

To give the reader an extensive overview of the tumor and error distributions, we created a qualitative distribution map of the top five teams in Fig. \ref{fig_errors2}.  
The data distribution shift is clearly visible in the reference annotations. While \ac{fdg} datasets show much uptake concentrated in the lungs, \ac{psma} datasets have a lot of activation in the bones (ribs, spine, and hips) and prostate if not resected and close to the bladder. A large proportion of false positives are visible in the head region and extremities. The head region is immediately apparent in the \uktpsma dataset. Here, we identified lacrimal gland uptake as a systematic error source. This physiological uptake was not visible in the training data nor in the other test sets. The other glands in the head region also experienced elevated false-positive segmentations. \lmufdg showed apparent FP attention to the spine, most likely reflecting increased uptake due to degenerative changes.

\subsubsection{Removal of lacrimal glands}
To verify the influence of the simple unseen physiology, we removed roughly 50 voxels from the top of each \uktpsma case (no reference annotation was removed) and recalculated the \ac{dsc} and \ac{fpv} for each case and team. Table \ref{tab:alternative_metrics} shows the updated average scores. Three observations become apparent. First, algorithms improve by different amounts. Those that already performed well in \uktpsma gained almost nothing (IKIM B: +0.02\%, AiraMatrix A: +0.07\%), while LesionTracer A gained nearly 5\% \ac{dsc} and Zero Sugar A a staggering 11.5\%. 
Second, the \ac{fpv} remains present, and algorithms that performed better on the original \uktpsma data still tend to have a lower \ac{fpv} than those that only become comparable in \ac{dsc} after cropping. Third, the predicted volume remains systematically higher than the reference volume across nearly all methods, even after removing the unseen region. We want to note, however, that by masking the lacrimal glands, we may also have removed other small false positives in the upper extremities, as visible in Figure \ref{fig_errors}.

\section{Discussions and conclusion}
The \ap challenge evaluated automated whole-body PET/CT lesion segmentation in a compositional generalization setting with two tracers and centers. With 17 final teams submitting 27 publicly available algorithms, the challenge yielded several findings. We structure the following discussion into methodological findings, performance, and biomedical findings, and challenge validity and metrics. 

\subsection{Methodological findings}
The submitted algorithms showed methodological homogeneity, consistent with the findings of previous challenges \citep{oreiller_head_2022, gatidis_results_2024} and many related studies (Sec.~\ref{sec:relatedalgo}). Twelve of 15 \ac{ac1} teams used nnU-Net, and all concatenated PET and CT as a multi-channel input. 
Two paradigms for multi-tracer handling emerged: single global models (11 teams) and two-stream routing with tracer-specific expert networks (4 teams). The winning algorithm (LesionTracer A) used a global model, but tracer-expert approaches (e.g., IKIM) achieved near-identical performance, questioning the benefit of multitracer learning. Tracer differentiation between \ac{psma} and \ac{fdg} PET/CT samples can be considered a well-solved problem, with participants using small discriminator models on MIPs (Sec. \ref{sec:multistage}).

This convergence raises the question of what actually differentiates performance among these architecturally similar approaches. The gap between the data-centric baseline and the top-performing algorithms is primarily driven by a reduction in \ac{fnv} (Table \ref{tab:alternative_metrics}).
However, no single design choice can be identified as the decisive factor. Most teams combined multiple modifications regarding backbone size, FOV, training parameters, sampling strategy, organ masks, and pre-processing, making it difficult to attribute the performance gain to any individual change.
Some trends are visible but not consistent (Sec. \ref{sec:algo-overview}). Backbone size relative to the baseline seems to be the most influential factor, which participants also confirmed through cross-validation experiments. Some top-ranked teams adopted upscaled backbone variants with a field of view around $192^3$ voxels, yet teams employing even larger backbones and FOV did not rank higher overall. The largest models achieved the lowest \ac{fpv}, suggesting that increased capacity may help suppress false positives from physiological uptake, though, in the case of AiraMatrix, body cropping that removes the head could equally explain the reduction in \ac{fpv}. Expert models on the other hand used a smaller FOV and smaller backbones. Anatomical label priors showed no clear benefit either: they were used by both top-and mid-ranked teams. In contrast, the third-ranked team used a plain nnU-Net differing from the others primarily in training exclusively on tumor-positive cases and using a larger batch size.

The data-centric track (\ac{ac2}) had, unfortunately, few participants, and the submitted strategies were highly heterogeneous. The winning solutions performed only slightly better than the baseline itself. Surprisingly, the misalignment strategy, which benefited the model for the winning team in \ac{ac1}, even performed worse than the data-centric baseline. This highlights that new methods sometimes do not translate to slightly different model configurations. MICCAI session participants expressed strong interest in this category; however, they noted that migrating from the nnU-Net ecosystem to the required MONAI implementation posed a significant barrier. The constrained FOV and backbone, chosen to ensure accessibility across hardware setups, may have limited the potential of data-centric strategies. Future work should investigate whether the observed trends hold with larger model configurations. Despite these caveats, we still think it is valuable to pursue data-centric approaches. Major hurdles will be disentangling effects in a fair and non-limited challenge setting.


\subsection{Performance and biomedical findings}
Whether the task of automated lesion segmentation in PET/CT is solved requires a nuanced answer. On in-domain datasets, top algorithms produce clinically relevant segmentations for the majority of cases (Fig.~\ref{fig_errors}). Lesion detectability is high, absolute volume differences are symmetric, and in \uktfdg almost all relative volume differences are within ±20\% for cases with a volume larger than 1 mL. Performance on \lmupsma is slightly lower, mainly due to undersegmentation and higher \ac{fnv}. The median \ac{dsc} is slowly approaching or exceeding approximate reader variability, though more rigorous reader studies are needed to confirm this (Sec.~\ref{sec:reader}). By introducing an open \ac{psma} dataset, we enabled the community to develop algorithms that segment \ac{psma} cases with substantially better performance compared to the previous challenge iteration \citep{dexl_autopet_2025}, largely driven by a reduction in \ac{fpv} alongside similar or slightly improved detection sensitivity. An exploratory experiment across \lmupsma cases indicated that performance does not differ between \ac{psma} ligands, though wide confidence intervals should be considered when interpreting this result (Sec.~\ref{sec:ligands}). Importantly, multitracer generalization did not compromise \ac{fdg} performance when using a single model. 

The picture becomes more complex for the out-of-domain composites, which reveal two distinct failure modes. \lmufdg had a higher \ac{fnv} and lower \ac{fpv} than \uktpsma, yet still oversegmented more drastically overall. This suggests that \lmufdg tends to miss small lesions entirely while oversegmenting the ones it does detect, possibly because lesion size and uptake distributions differ from training (Sec.~\ref{sec:lesionsuv}). \uktpsma, by contrast, generates false positives from rare or unseen physiological patterns, the clearest example being lacrimal glands, which were nearly absent from training data (due to anonymization). Even after removing the head region to account for this, elevated \ac{fpv} persisted and varied considerably across teams, confirming that out-of-domain generalization is not straightforward, even under a much weaker compositional setting. 

A closer look at the error types reveals distinct patterns across both settings consistent with related works \citep{weikert_automated_2023, andrearczyk_automatic_2023, park_automatic_2023}. 
Most false negatives appear in lesions that are also challenging for physicians: small lesions with low uptake, or cases with a single small lesion (Sec. \ref{sec:lesionsuv}). In clinical practice, physicians typically have access to additional context, such as the patient history, laboratory reports, or a second opinion. Some false negatives, however, are better explained by rarity in the training data. We also note that lesions are often confirmed on CT, and some lesions with low SUV uptake are clearly visible on CT. Disentangling the extent to which models rely on CT versus PET information would be a worthwhile investigation, but it was beyond the scope of this work. 

False positives remain the primary area for improvement and fall into two categories. The first, consists of annotation errors in the ground-truth labels: algorithms detected lesions that were missed during annotation, particularly in high-tumor-burden cases with extensive bone metastases, where exhaustive labeling is exceptionally time-consuming. The second, more concerning category comprises false positives that would be obvious to a physician, such as tracer contamination at injection sites, muscle activation, or physiological uptake in glands or the bladder. Notably, the post-hoc majority-vote ensemble of the top five algorithms substantially reduces \ac{fpv} while leaving \ac{dsc} and \ac{fnv} largely unchanged (Sec. \ref{sec:ensembling}), suggesting that these false positives are spatially diverse across models and therefore not an inherent limitation of the task.

Among the top-18 submissions, the choice of case matters far more than the choice of algorithm. Patient heterogeneity accounts for 61\% of the unexplained variance in DSC, compared to only 1.3\% attributable to team differences, a roughly seven-fold difference in standard deviations (Sec. \ref{sec:mixedmodel}). 
In addition, almost all metric distributions are non-normal, with long tails of difficult cases that challenge algorithms and physicians alike. The mean consistently underestimates typical performance, which explains a common observation when presenting segmentations to physicians: the majority of cases look excellent, yet aggregate metrics appear modest.
Clinically, this shifts the priority from algorithm selection to case triage: difficult cases should be identified upfront, and physician verification remains essential for this tail.

A related limitation emerges when including lesion-free cases: current models should not be used for binary classification (Sec. \ref{sec:classification}). Many algorithms were overly sensitive and flagged uptake even in negative cases, though the associated volumes were typically small. This is plausible, as many of these patients are post-treatment, where therapy-related changes (e.g., post-chemotherapy or post-surgical effects) can produce suspicious-looking uptake patterns.

\subsection{Challenge validity and metrics}
Do we trust the challenge results and rankings? The bootstrap analysis (Fig.~\ref{fig_ranking_stability}) and the alternative ranking methods (Fig.~\ref{fig:alternative_rankings}) reveal an overall two-tier structure. In the better tier, there still seems to be a structural ordering, i.e., teams with higher performance ranking higher more often across methods and bootstraps. 

We observe two flaws in our ranking that are worth noting. First, we ranked within each subgroup, which neglected absolute performance differences. Here, the weighted-average ranking (R2) might be more suitable; the ranking change, however, is subtle (Sec. \ref{sec:altrank}). Second, the composite metric is biased towards sensitivity. Lesion-free patients contribute only through \ac{fpv}, while \ac{dsc} (weighted double) and \ac{fnv} are computed exclusively on lesion-positive cases. This incentivizes small positive volumes and undervalues algorithms with great \ac{fpv} performance. 

Throughout the autoPET editions, there has been extensive discussion about the challenge metrics. Lesion segmentation can also be interpreted as instance segmentation, and there has been recurring criticism of the one-voxel detection criterion used for \ac{fnv} and \ac{fpv} computation. The concern is that imprecise predictions could disproportionately benefit algorithms in the challenge ranking, or that the metrics are hackable, for example, by connecting all lesions. 
What does the lesion-level analysis teach us in this regard?
Our lesion-level analysis shows that much of the ambiguity lies not in the metric but in the reference annotations themselves. Lesion boundaries in PET/CT are inherently fuzzy and often fragmented due to threshold-based labeling or CT-derived contours with small spatial misalignments. In many cases, a single voxel connects or divides two segments. Even the connectivity criterion influences the reference lesion count. We believe that as long as an acceptable segmentation quality is ensured, \ac{fpv} and \ac{fnv} remain reasonable metrics. When applying a stricter detection criterion, although median sensitivity drops roughly by 35\%, it seems that most algorithms are stable to this shift (better algorithms stay better) (Fig. \ref{fig_lesion_dect}). Based on the error analysis, we can now precisely quantify the associated error types and their variances (Sec. \ref{sec:lesionerr}). Predictions covering multiple lesions are slightly more common than reference annotations split into multiple segmentations at low thresholds. At the one-voxel criterion, roughly 20\% of the total lesions are in clusters.

This discussion of the evaluation metrics is ultimately connected to a broader, unresolved problem: the absence of clinical importance weighting for individual lesions. The current \ac{fnv}/\ac{fpv} logic implicitly weighs lesions by size, which is reasonable but incomplete. Count-based detection metrics, on the other hand, weigh each error equally. In reality, a small, distant metastasis that upstages a patient from oligometastatic to disseminated disease can carry more weight in a treatment decision than a large, known primary tumor. Incorporating location-dependent weighting would better align the metrics with clinical decision-making, but it would require domain experts to define region-specific importance weights, which is a non-trivial task. A simpler first step might be to categorize cases into those where detection matters and those where it does not, and to adapt the evaluation accordingly. 

\subsection{Limitations}
Several limitations of the challenge design should be acknowledged. First, each tracer–center condition coincides with a unique combination of scanner hardware, reconstruction protocol, annotation method, and patient demographics. These factors are, in part, confounded with the condition labels, so the fixed effects in the mixed-effects model cannot isolate a single source of variation, and their estimates should be interpreted accordingly.

Second, while the five-minute runtime constraint per case can be viewed as beneficial by ensuring models are deployable within a reasonable time, it may result in limited performance. Participants also reported this constraint as difficult to manage. More computationally demanding approaches, including heavy ensembling, full test-time augmentation, and larger models, were all affected, and the ranking may therefore not fully reflect what is achievable without this constraint.

Third, the reader comparison should be interpreted as an approximate reference point. Reading protocols (inter vs intra) and reader experience differed across conditions, sample sizes were small (n = 25 per condition), and no second-reader data were available for \uktpsma. 

Fourth, possible errors in the ground-truth labels and an annotation shift exist between datasets: \uktfdg lesions were delineated slice-by-slice, while the others used threshold-based pre-segmentation with manual refinement, which may introduce systematic boundary differences. Also, reading was done by a single annotator, which introduces a label bias.

\subsection{Looking forward and concluding thoughts}
The \ap challenge demonstrates that multitracer, multicenter PET/CT lesion segmentation is feasible, with good in-domain performance that likely approaches reader agreement. Compositional generalization, i.e., recombining knowledge of tracers and centers in unseen combinations, is achievable but remains constrained by systematic and distributional errors, which can only be addressed by expanding the training distribution or making smart prior assumptions. Based on our findings, we identify three priorities for the field. 

First, simply another challenge edition with similar nnU-Net based submissions will not advance our understanding. The error analysis reveals persistent obvious false positives, suggesting it is time to shift toward interactive segmentation approaches that allow clinicians to correct outlier cases. AutoPET IV will take this as its primary focus.

Second, dataset quality and heterogeneity remain the largest bottleneck, especially since autoPET aims to generalize across multiple disease types and tracers. Future editions should expand the tracer and center landscape (e.g., [\textsuperscript{18}F]FAPI, [\textsuperscript{68}Ga]DOTATATE) and prioritize case variance, difficult examples, and harmonized annotations over sheer dataset size. Iterative algorithm-assisted labeling may help offset annotation costs.

Third, our metrics are only surrogates that must be translated into clinical utility. In this context, several clinical questions are increasingly important: What volume and detection error rates are acceptable, which lesions are most relevant, and how do requirements differ across disease types and tracers? Nevertheless, we believe that current algorithms already produce valuable results that can support research applications and expert-supervised clinical workflows, representing a meaningful step toward routine clinical translation.

\section*{CRediT authorship contribution statement}
\textbf{Jakob Dexl:} Conceptualization, Data curation, Formal analysis, Investigation, Methodology, Project administration, Resources, Software, Validation, Visualization, Writing -- original draft, Writing -- review \& editing. \textbf{Katharina Jeblick:} Conceptualization, Data curation, Project administration, Resources, Software,  Writing -- review \& editing. \textbf{Andreas Mittermeier:} Conceptualization, Data curation, Project administration, Resources, Software,  Writing -- review \& editing. \textbf{Balthasar Schachtner:} Data curation, Resources, Software, Writing -- review \& editing. \textbf{Anna Theresa Stüber:} Conceptualization, Formal analysis, Resources, Software, Validation, Writing -- review \& editing. \textbf{Johanna
Topalis:} Writing -- review \& editing. \textbf{Maximilian Rokuss:} Methodology, Resources, Software, Writing -- review \& editing. \textbf{Fabian Isensee:} Methodology, Resources, Software, Writing -- review \& editing., \textbf{Klaus H. Maier-Hein:} Methodology, Resources, Software, Writing -- review \& editing. \textbf{Hamza Kalisch:} Methodology, Resources, Software, Writing -- review \& editing. \textbf{Jens Kleesiek:} Methodology, Resources, Software, Writing -- review \& editing. \textbf{Constantin M.
Seibold:} Methodology, Resources, Software, Writing -- review \& editing. \textbf{Hussain Alasmawi:} Methodology, Resources, Software, Writing -- review \& editing. \textbf{Lap Yan Lennon Chan:} Methodology, Resources, Software, Writing -- review \& editing. \textbf{Yixuan Yuan:} Methodology, Resources, Software, Writing -- review \& editing. \textbf{Alexander Jaus:} Methodology, Resources, Software, Writing -- review \& editing. \textbf{Rainer Stiefelhagen:} Methodology, Resources, Software, Writing -- review \& editing. \textbf{Pauline Ornela
Megne Choudja:} Writing -- review \& editing. \textbf{Konstantin Nikolaou:} Writing -- review \& editing. \textbf{Christian La Fougère:} Data Curation, Writing -- review \& editing. \textbf{Sergios Gatidis:} Data Curation, Writing -- review \& editing. \textbf{Matthias P. Fabritius:} Data curation, Validation, Writing -- review \& editing.  \textbf{Maurice Heimer:} Validation, Writing -- review \& editing. 
\textbf{Gizem Abaci:} Data curation, Writing -- review \& editing. \textbf{Lalith Kumar Shiyam Sundar:} Writing -- review \& editing. \textbf{Rudolf A. Werner:} Data Curation, Writing -- review \& editing. \textbf{Jens Ricke:} Funding acquisition, Resources, Writing -- review \& editing. \textbf{Clemens C. Cyran:} Funding acquisition, Conceptualization, Data curation, Project administration, Resources, Supervision, Writing -- review \& editing. \textbf{Thomas Küstner:} Conceptualization, Data curation, Funding acquisition, Project administration, Resources, Supervision, Writing -- review \& editing. \textbf{Michael
Ingrisch:} Conceptualization, Data curation, Funding acquisition, Project administration, Resources, Supervision, Writing -- review \& editing.

\section*{Acknowledgments}
We thank all the participants of the \ap challenge for their contributions. 

The authors gratefully acknowledge the LMU University Hospital for providing computing resources on their Clinical Open Research Engine (CORE). 

This paper is supported by the DAAD program Konrad Zuse Schools of Excellence in Artificial Intelligence, sponsored by the Federal Ministry of Research, Technology, and Space. This project was conducted under Germany's Excellence Strategy EXC-Numbers EXC 2064/1-390727645 and EXC 2180/1-390900677. 

This work was supported in part by the Cluster of Excellence iFIT (EXC 2180) "Image Guided and Functionally Instructed Tumor Therapies", German Research Foundation – Excellence Initiative.

Katharian Jeblick was supported through funding from Bayerisches Staatsministerium für Wissenschaft und Kunst in cooperation with Fonds de Recherche Santé Québec and by the Protected Time 4 Research Programme 2025 – Special BGF programme for female postdoctoral researchers at LMU.

Lap Yan Lennon Chan was supported through a research grant from the Faculty of Engineering of the Chinese University of Hong Kong for Undergraduate Summer Research Internship programme 2024.

\section*{Declaration of competing interest}
Rudolf A. Werner reports a relationship with Novartis that includes: speaking and lecture fees. The other authors, declare that they have no known competing financial interests or personal relationships that could have appeared to influence the work reported in this paper.

\section*{Data availability}
Data are publicly available.

\section*{Declaration of generative AI and AI-assisted technologies in the manuscript preparation process}
During the drafting of the manuscript, Opus 4.6 from Anthropic, GPT 5.2 from OpenAI and Grammarly were utilized to enhance language, clarity and structure. Additionally, Opus 4.6 was used as a programming companion for refining analysis and visualization code. After using these tools/services, the author(s) reviewed and edited the content as needed and take(s) full responsibility for the content of the published article.

\appendix

\section{Top performing teams}
\label{sec:topalgo}
\noindent\textbf{LesionTracer (\ac{ac1}):} \citep{rokuss_fdg_2024} The approach is based on the nnU-Net framework with a ResEnc L backbone and employs a dual-headed design, with one head for lesion segmentation and one for organ segmentation. Training was performed on 3D patches of size ($192\times192\times192$). PET volumes are normalized using a global normalization. Dice loss without smoothing is applied equally to both heads. Data augmentation includes standard nnU-Net transforms and a novel misalignment augmentation that simulates PET/CT registration errors. The model was pretrained on a large multimodal dataset combining CT, MR, and PET images similar to \citet{ulrich_multitalent_2023}, and subsequently fine-tuned on the challenge data. The submitted model (LesionTracer A) was trained with the Adam optimizer and a batch size of 3 for 1,500 epochs and uses a 5-fold ensemble with a larger tile step size. Depending on the remaining time, mirroring-based test-time augmentation is applied after inference on the second to fifth folds. The authors conducted several ablations and concluded that increasing the backbone yielded the largest gains, and that isotropic resampling to a spacing of 1\,mm, pretraining on organ masks from TotalSegmentator, or the use of additional \ac{fdg} data from HECKTOR did not enhance performance. The team submitted a second algorithm to \ac{ac2} employing the misalignment augmentation (LesionTracer B).\\

\noindent\textbf{IKIM (\ac{ac1}):} \citep{kalisch_autopet_2024}
The method employs a multi-stage approach utilizing a tracer-classification model and tracer-specific nnU-Nets for lesion segmentation. First, coronal and sagittal Maximum Intensity Projections (MIP) of the PET volume are passed through two ResNet18 \citep{he_deep_2016} backbones. Their frozen feature vectors are fused via a multi-layer perceptron to predict the tracer. Based on this classification, a dedicated nnU-Net with a ResEnc M encoder and an input size of ($128\times112\times160$) (FDG) or ($96\times112\times224$) (PSMA) processes the concatenated PET/CT volumes. Organ masks generated by the TotalSegmentator are incorporated as auxiliary segmentation targets alongside lesion labels, with a weighting factor in the Dice-CE loss balancing the anatomy and lesion objectives. Each tracer model uses different organ subsets. PET volumes are normalized per sample. The models were trained with default nnU-Net augmentations and a batch size of 2 for 1,000–1,500 epochs. For post-processing, segmentations are thresholded based on SUV values (1.5 for FDG, 1.0 for PSMA). The team submitted two algorithms to the challenge, both using the same FDG model weights paired with different PSMA model runs. The authors also experimented with fine-tuning and connected-component thresholding as post-processing; however, both were discarded. \\

\noindent\textbf{HussainAlasmawi (\ac{ac1}):} \citep{alasmawi2024autopet} The approach uses a vanilla nnU-Net with a ResEnc L backbone and a patch size of ($192\times192\times192$), trained exclusively on patients with tumors. The team submitted two algorithms that differ in PET normalization and loss aggregation. The first (HussainAlasmawi B) was trained with z-score normalization for PET volumes. The second (HussainAlasmawi A) used global normalization and adopted a batch-level dice loss, inspired by \citet{isensee_nnu-net_2021-1}, combined with an increased batch size of 5. The authors also experimented with the CarveMix augmentation \citep{zhang_carvemix_2023} and added 350 synthetic cases; improvements were moderate, and due to the submission limit, this method was not used. \\

\noindent\textbf{Lennonlychan (\ac{ac2}):} \citep{li_automated_2024} This data-centric approach adapts the DiffTumor \citep{chen_towards_2024} pipeline from CT-only to paired PET/CT synthesis for data augmentation. Only the first two stages of DiffTumor are used: an autoencoder is first trained on AutoPET PET/CT samples to learn a compressed latent space, then a latent diffusion model is trained on the full AutoPET dataset to generate tumorous PET/CT latents conditioned on lesion and organ masks. Organ masks are predicted by a MONAI SegResNet model \citep{myronenko_3d_2019} trained on the TotalSegmentor dataset. Finally, three samples are generated per training case. The fixed baseline is then trained with a batch size of 2 for 581 epochs. \\

\noindent\textbf{ZeroSugar (\ac{ac2}):} \citep{jaus_data_2024} This data-centric approach is based on a pruning strategy inspired by recent work on dataset filtering \citep{gadre_datacomp_2023} to counteract systematic overconfidence observed in \ac{psma} PET volumes. The fixed model is used to compute the per-sample loss, \ac{dsc}, \ac{fpv}, and \ac{fnv} across all training cases. Analysis reveals that \ac{psma} studies exhibit a distinct right-shift in \ac{fpv} and a pronounced imbalance in cases with and without lesions compared to \ac{fdg}. To mitigate this, \ac{psma} samples were sorted by baseline loss in ascending order, and the lowest-loss percentile was excluded from training; \ac{fdg} samples were retained in full. This removes overly easy \ac{psma} cases, none of which contain healthy patients, and preserves harder examples for improved calibration. Excluding the 3rd percentile of easiest PSMA volumes yielded the highest \ac{dsc}, while excluding the 5th percentile yielded the best \ac{fnv}.

\section{Mixed-effects model specifications}
\label{sec:apmixed}
All linear mixed models were fitted using \texttt{lme4} in R with REML. Model formulas use R notation. Approximate 95\% CIs are profile-type.\\

\noindent Model A1 (\ac{dsc}):
\\
\texttt{dsc $\sim$ tracer * center + $\log_2$(V) + (1 $|$ team) + (1 $|$ patient)}
\begin{minipage}[t]{0.48\textwidth}
\captionsetup{justification=raggedright, singlelinecheck=false}
\centering
\small
\captionof{table}{Fixed effects \ac{dsc} model.}
\begin{tabular}{lrrl}
\hline
 & $\beta$ & SE & 95\% CI \\
\hline
Intercept            &  0.410 & 0.036 & $[0.341,\; 0.479]$ \\
Tracer (\ac{psma})        & $-$0.037 & 0.039 & $[-0.113,\; 0.039]$ \\
Center (UKT)         &  0.140 & 0.040 & $[0.063,\; 0.217]$ \\
$\log_2(V)$          &  0.039 & 0.005 & $[0.030,\; 0.048]$ \\
Tracer$\times$Center & $-$0.081 & 0.057 & $[-0.193,\; 0.030]$ \\
\hline
\end{tabular}
\end{minipage}
\hfill
\begin{minipage}[t]{0.48\textwidth}
\captionsetup{justification=raggedright, singlelinecheck=false}
\end{minipage}

\noindent Random-effect standard deviations: patient intercept $\mathrm{SD} = 0.171$, team intercept $\mathrm{SD} = 0.025$, residual $\mathrm{SD} = 0.134$.
$n_\text{obs} = 2{,}808$; $n_\text{patient} = 156$; $n_\text{team} = 18$. REML criterion: $-2{,}714.8$. Scaled residuals: $[-5.02,\; 4.96]$. Residual diagnostics show an S-shaped QQ-plot with leptokurtic tails and heteroscedasticity that compresses near both bounds of DSC. Random-effect quantiles for both grouping factors appear approximately normal.
\bigskip

\noindent Model 2 (Radionuclide effect, \ac{psma}--LMU subset):
\texttt{dsc $\sim$ radionuclide + (1 $|$ patient) + (1 $|$ team)}
\begin{minipage}[t]{0.48\textwidth}
\centering
\small
\captionof{table}{Fixed effects \ac{psma} ligand model.}
\begin{tabular}{lrrl}
\hline
 & $\beta$ & SE & 95\% CI \\
\hline
Intercept                    &  0.576 & 0.067 & $[0.446,\; 0.706]$ \\
$^{68}$Ga vs.\ $^{18}$F     & $-$0.027 & 0.095 & $[-0.213,\; 0.159]$ \\
\hline
\end{tabular}
\end{minipage}
\hfill
\begin{minipage}[t]{0.48\textwidth}
\small
\end{minipage}
\newline
\noindent Random-effect standard deviations: patient intercept $\mathrm{SD} = 0.272$, team intercept $\mathrm{SD} = 0.030$, residual $\mathrm{SD} = 0.110$.
$n_\text{obs} = 594$; $n_\text{patient} = 33$; $n_\text{team} = 18$. REML criterion: $-760.0$. Scaled residuals: $[-4.72,\; 6.03]$. Residual diagnostics look similar to the DSC model. 
\bigskip

\section{Additional tables and figures}
See Tables \ref{tab:raw_results}, \ref{tab:alternative_metrics}, and Figures \ref{fig:individual_metrics} to \ref{fig:suv_uncert}.

\begin{landscape}
\begin{table}[h]
\centering
\caption{Official results of all submitted algorithms. For each challenge metric, results are reported separately per dataset, with ranks shown in parentheses. The final position is determined by combining the per-metric ranks; only the best-performing submission per team is considered for positioning. Bold values indicate best performance per column. Methods marked with * denote reference methods and the post-hoc ensembles. The bottom three algorithms were not taken into account for ranking.}
\def\arraystretch{1.5}
\resizebox{\linewidth}{!}{%
\begin{tabular}{@{}
>{\columncolor[HTML]{FFFFFF}}l 
>{\columncolor[HTML]{FFFFFF}}c 
>{\columncolor[HTML]{FFFFFF}}l 
>{\columncolor[HTML]{FFFFFF}}l 
>{\columncolor[HTML]{FFFFFF}}l 
>{\columncolor[HTML]{FFFFFF}}l 
>{\columncolor[HTML]{FFFFFF}}l 
>{\columncolor[HTML]{FFFFFF}}l 
>{\columncolor[HTML]{FFFFFF}}l 
>{\columncolor[HTML]{FFFFFF}}l 
>{\columncolor[HTML]{FFFFFF}}l 
>{\columncolor[HTML]{FFFFFF}}l 
>{\columncolor[HTML]{FFFFFF}}l 
>{\columncolor[HTML]{FFFFFF}}l 
>{\columncolor[HTML]{FFFFFF}}l 
>{\columncolor[HTML]{FFFFFF}}l 
>{\columncolor[HTML]{FFFFFF}}l 
>{\columncolor[HTML]{FFFFFF}}l @{}}
\toprule
   Team &
   Position &
   Rank &
  \multicolumn{5}{l}{Dice Similarity Coefficient} &
  \multicolumn{5}{l}{False Negative Volume (mL)}  &
  \multicolumn{5}{l}{False Positive Volume (mL)}
   \\
\midrule &
   &
   &
  \begin{tabular}[c]{@{}l@{}}\uktfdg\end{tabular} &
  \begin{tabular}[c]{@{}l@{}}\lmupsma\end{tabular} &
  \begin{tabular}[c]{@{}l@{}}\lmufdg\end{tabular} &
  \begin{tabular}[c]{@{}l@{}}\uktpsma\end{tabular} &
  \begin{tabular}[c]{@{}l@{}}Rank\end{tabular} &
  \begin{tabular}[c]{@{}l@{}}\uktfdg\end{tabular} &
  \begin{tabular}[c]{@{}l@{}}\lmupsma\end{tabular} &
  \begin{tabular}[c]{@{}l@{}}\lmufdg\end{tabular} &
  \begin{tabular}[c]{@{}l@{}}\uktpsma\end{tabular} &
  \begin{tabular}[c]{@{}l@{}}Rank\end{tabular} &
  \begin{tabular}[c]{@{}l@{}}\uktfdg\end{tabular} &
  \begin{tabular}[c]{@{}l@{}}\lmupsma\end{tabular} &
  \begin{tabular}[c]{@{}l@{}}\lmufdg\end{tabular} &
  \begin{tabular}[c]{@{}l@{}}\uktpsma\end{tabular} &
  \begin{tabular}[c]{@{}l@{}}Rank\end{tabular} \\
\hline
LesionTracer A &
  1 &
  6.0625 &
  0.7702 (7) &
  \textbf{0.6433} (1) &
  0.6619 (3) &
  0.5711 (8) &
  \textbf{4.7500} &
  \textbf{0.4310} (1) &
  10.1590 (8) &
  \textbf{1.7810} (1) &
  \textbf{0.3658} (1) &
  \textbf{2.7500} &
  3.0189 (23) &
  1.1123 (9) &
  1.7030 (5) &
  5.3053 (11) &
  12.0000 \\
IKIM A &
  2 &
  8.4062 &
  0.7938 (3) &
  0.5828 (7) &
  0.6043 (16) &
  0.6176 (4) &
  7.5000 &
  1.0422 (8.5) &
  8.4799 (2) &
  2.7249 (3) &
  0.7334 (7) &
  5.1250 &
  2.6586 (19) &
  1.4709 (14) &
  4.9373 (17) &
  3.1085 (4) &
  13.5000 \\
HussainAlasmawi A &
  3 &
  8.4375 &
  0.7802 (5) &
  0.5725 (10) &
  0.6370 (6) &
  0.5854 (7) &
  7.0000 &
  0.6365 (2) &
  10.1997 (9) &
  2.4542 (2) &
  0.6265 (4) &
  4.2500 &
  2.8756 (22) &
  1.7221 (18) &
  3.3616 (13) &
  5.2173 (9) &
  15.5000 \\
IKIM B &
  - &
  8.6562 &
  0.7938 (2) &
  0.5698 (11) &
  0.5900 (18) &
  \textbf{0.6642} (1) &
  8.0000 &
  1.0422 (8.5) &
  10.3443 (10) &
  2.8941 (4) &
  0.8312 (9) &
  7.8750 &
  2.6591 (20) &
  0.9713 (7) &
  4.7393 (15) &
  \textbf{1.1396} (1) &
  10.7500 \\
HussainAlasmawi B &
  - &
  9.8750 &
  \textbf{0.8105} (1) &
  0.6245 (2) &
  0.5756 (22) &
  0.5661 (9) &
  8.5000 &
  0.6746 (5) &
  9.9361 (6) &
  8.3681 (10) &
  0.7257 (6) &
  6.7500 &
  1.8248 (13) &
  1.4965 (15) &
  6.3350 (19) &
  6.7886 (16) &
  15.7500 \\
StockholmTrio &
  4 &
  9.9375 &
  0.7684 (8) &
  0.6010 (4) &
  0.5771 (21) &
  0.6326 (3) &
  9.0000 &
  1.1839 (12) &
  11.1818 (19) &
  8.4921 (11) &
  0.8581 (11) &
  13.2500 &
  1.7990 (12) &
  0.8850 (6) &
  1.9526 (9) &
  3.8848 (7) &
  8.5000 \\
UIH-CRI-SIL A &
  5 &
  10.0625 &
  0.7922 (4) &
  0.5937 (5) &
  0.6194 (14) &
  0.4111 (18) &
  10.2500 &
  0.9560 (7) &
  \textbf{7.5758} (1) &
  4.6745 (6) &
  0.4422 (2) &
  4.0000 &
  2.2770 (15) &
  2.0641 (21) &
  1.9500 (8) &
  12.4480 (19) &
  15.7500 \\
AiraMatrix A &
  6 &
  11.0625 &
  0.7556 (10) &
  0.5273 (20) &
  0.6253 (11) &
  0.6600 (2) &
  10.7500 &
  1.6105 (16) &
  15.6801 (28) &
  12.5127 (17) &
  1.0412 (16) &
  19.2500 &
  0.8604 (5) &
  \textbf{0.4382} (1) &
  1.8776 (6) &
  1.7253 (2) &
  \textbf{3.5000} \\
Lennonlychan A &
  7 &
  11.6250 &
  0.7454 (13) &
  0.5302 (18) &
  0.6286 (9) &
  0.4775 (10) &
  12.5000 &
  1.1815 (11) &
  10.5936 (11) &
  12.5980 (18) &
  0.8377 (10) &
  12.5000 &
  1.2379 (7) &
  1.0702 (8) &
  1.8980 (7) &
  5.9060 (14) &
  9.0000 \\
QuantIF A &
  8 &
  11.6875 &
  0.7649 (9) &
  0.5470 (13) &
  0.5800 (20) &
  0.6024 (6) &
  12.0000 &
  2.2904 (20) &
  11.7130 (20) &
  6.7282 (8) &
  0.8140 (8) &
  14.0000 &
  \textbf{0.7074} (1) &
  0.6253 (3) &
  10.0465 (23) &
  4.2608 (8) &
  8.7500 \\
QuantIF B &
  - &
  11.7500 &
  0.7731 (6) &
  0.5332 (17) &
  0.5841 (19) &
  0.6172 (5) &
  11.7500 &
  2.0833 (19) &
  12.0557 (22) &
  6.5790 (7) &
  0.9153 (13) &
  15.2500 &
  0.7258 (2) &
  0.6749 (4) &
  7.7236 (21) &
  3.5425 (6) &
  8.2500 \\
Shadab &
  9 &
  12.0625 &
  0.7270 (16) &
  0.5794 (9) &
  0.6159 (15) &
  0.4540 (13) &
  13.2500 &
  0.7998 (6) &
  9.9943 (7) &
  11.4725 (15) &
  0.6623 (5) &
  8.2500 &
  1.9246 (14) &
  1.4117 (13) &
  2.1096 (10) &
  7.0941 (17) &
  13.5000 \\
ZeroSugar A &
  10 &
  13.2500 &
  0.7373 (15) &
  0.5446 (14) &
  0.6580 (4) &
  0.4271 (17) &
  12.5000 &
  4.3788 (25) &
  11.0423 (17) &
  12.3233 (16) &
  1.6138 (22) &
  20.0000 &
  0.8209 (4) &
  1.3717 (11) &
  0.5802 (2) &
  6.3086 (15) &
  8.0000 \\
BAMF A &
  11 &
  13.3125 &
  0.6973 (22) &
  0.5299 (19) &
  \textbf{0.6666} (1) &
  0.4672 (11) &
  13.2500 &
  1.4420 (15) &
  14.7570 (26) &
  11.3668 (14) &
  2.3205 (26) &
  20.2500 &
  1.7017 (10) &
  0.6220 (2) &
  3.1945 (11) &
  2.2677 (3) &
  6.5000 \\
BAMF B &
  - &
  13.5000 &
  0.6912 (24) &
  0.5642 (12) &
  0.6664 (2) &
  0.4384 (16) &
  13.5000 &
  1.3328 (13) &
  13.0037 (25) &
  7.3285 (9) &
  2.6724 (28) &
  18.7500 &
  1.6624 (9) &
  0.7829 (5) &
  4.0591 (14) &
  3.4041 (5) &
  8.2500 \\
ZeroSugar B &
  - &
  14.1875 &
  0.7004 (20) &
  0.5364 (16) &
  0.6505 (5) &
  0.4524 (14) &
  13.7500 &
  3.9189 (23) &
  11.0095 (16) &
  18.3225 (24) &
  1.2524 (20) &
  20.7500 &
  0.8089 (3) &
  1.5535 (17) &
  0.8088 (4) &
  5.2175 (10) &
  8.5000 \\
Lennonlychan B &
  - &
  14.4375 &
  0.7410 (14) &
  0.5255 (21) &
  0.6351 (8) &
  0.4558 (12) &
  13.7500 &
  1.7714 (18) &
  10.5945 (12) &
  16.5714 (22) &
  0.8729 (12) &
  16.0000 &
  2.5750 (17) &
  1.4967 (16) &
  3.2294 (12) &
  5.5124 (12) &
  14.2500 \\
*datacentric baseline &
  - &
  15.3125 &
  0.7254 (18) &
  0.5252 (22) &
  0.6362 (7) &
  0.4397 (15) &
  15.5000 &
  5.0262 (26) &
  11.0583 (18) &
  15.6634 (21) &
  1.0951 (17) &
  20.5000 &
  0.9542 (6) &
  1.9186 (19) &
  \textbf{0.3659} (1) &
  5.7609 (13) &
  9.7500 \\
DING1122 &
  12 &
  18.7500 &
  0.7267 (17) &
  0.5390 (15) &
  0.6281 (10) &
  0.1062 (27) &
  17.2500 &
  1.3436 (14) &
  10.8470 (15) &
  14.0621 (19) &
  2.0637 (23) &
  17.7500 &
  2.3319 (16) &
  2.5160 (23) &
  11.2634 (25) &
  127.1480 (27) &
  22.7500 \\
WukongRT &
  13 &
  18.8750 &
  0.7221 (19) &
  0.5206 (23) &
  0.6204 (13) &
  0.3386 (22) &
  19.2500 &
  1.1040 (10) &
  11.7982 (21) &
  9.3683 (12) &
  1.0006 (15) &
  14.5000 &
  3.5422 (24) &
  3.3785 (25) &
  6.7212 (20) &
  18.7775 (21) &
  22.5000 \\
HKURad &
  14 &
  19.3750 &
  0.6973 (21) &
  0.4955 (25) &
  0.6026 (17) &
  0.3496 (21) &
  21.0000 &
  3.3258 (21) &
  12.9096 (24) &
  10.8671 (13) &
  1.1034 (18) &
  19.0000 &
  2.6552 (18) &
  1.3913 (12) &
  4.8969 (16) &
  13.1626 (20) &
  16.5000 \\
Maxsh A &
  15 &
  19.5312 &
  0.7483 (11.5) &
  0.5811 (8) &
  0.4012 (26) &
  0.0285 (28.5) &
  18.5000 &
  0.6570 (3.5) &
  9.2688 (3) &
  30.2820 (27) &
  2.1117 (24.5) &
  14.5000 &
  3.8532 (26) &
  3.0797 (24) &
  38.8093 (28) &
  924.3099 (28.5) &
  26.6250 \\
LesionTracer B &
  - &
  19.5625 &
  0.6110 (27) &
  0.4466 (28) &
  0.6218 (12) &
  0.3611 (19) &
  21.5000 &
  19.6584 (28) &
  14.7929 (27) &
  21.4829 (25) &
  1.2459 (19) &
  24.7500 &
  1.7559 (11) &
  1.1609 (10) &
  0.7751 (3) &
  10.6363 (18) &
  10.5000 \\
Maxsh B &
  - &
  19.7812 &
  0.7483 (11.5) &
  0.5834 (6) &
  0.3495 (27) &
  0.0285 (28.5) &
  18.2500 &
  0.6570 (3.5) &
  10.6836 (13) &
  38.6279 (29) &
  2.1117 (24.5) &
  17.5000 &
  3.8530 (25) &
  2.0140 (20) &
  29.8035 (27) &
  924.3099 (28.5) &
  25.1250 \\
AiraMatrix B &
  - &
  21.6250 &
  0.6972 (23) &
  0.5080 (24) &
  0.5284 (23) &
  0.2491 (24) &
  23.5000 &
  3.5834 (22) &
  10.7346 (14) &
  17.9929 (23) &
  1.5693 (21) &
  20.0000 &
  1.6278 (8) &
  2.2258 (22) &
  11.1968 (24) &
  37.2596 (24) &
  19.5000 \\
Shrajanbhandary &
  16 &
  21.7500 &
  0.1875 (29) &
  0.6125 (3) &
  0.2669 (29) &
  0.3597 (20) &
  20.2500 &
  54.5663 (29) &
  9.4384 (4) &
  32.2533 (28) &
  2.3478 (27) &
  22.0000 &
  13538 (29) &
  8429 (29) &
  5.4561 (18) &
  24.7666 (22) &
  24.5000 \\
UIH-CRI-SIL B &
  - &
  21.8750 &
  0.6469 (26) &
  0.4925 (26) &
  0.4298 (25) &
  0.2903 (23) &
  25.0000 &
  4.3688 (24) &
  9.5916 (5) &
  23.3383 (26) &
  0.5800 (3) &
  14.5000 &
  2.7256 (21) &
  4.5019 (26) &
  8.6082 (22) &
  30.6340 (23) &
  23.0000 \\
*nnUNet baseline &
  - &
  24.0000 &
  0.6822 (25) &
  0.4741 (27) &
  0.5017 (24) &
  0.1512 (26) &
  25.5000 &
  1.6795 (17) &
  12.5885 (23) &
  14.0836 (20) &
  0.9905 (14) &
  18.5000 &
  6.6433 (27) &
  4.9485 (27) &
  11.7733 (26) &
  102.6461 (26) &
  26.5000 \\
TUM-ibbm &
  17 &
  26.2500 &
  0.3639 (28) &
  0.2031 (29) &
  0.2845 (28) &
  0.1891 (25) &
  27.5000 &
  9.4993 (27) &
  28.4085 (29) &
  3.6257 (5) &
  3.4622 (29) &
  22.5000 &
  250.7309 (28) &
  216.4435 (28) &
  231.9891 (29) &
  89.9235 (25) &
  27.5000 \\
\bottomrule
  *nnUNet FDG only &
  - &
  - &
  0.8475 &
  0.1226 &
  0.5611 &
  0.0139 &
  - &
  3.0593 &
  25.2173 &
  3.4863 &
  5.4800 &
  - &
  1.1023 &
  910.0258 &
  7.4739 &
  1704.2898 &
  - \\
*nnUNet PSMA only &
  - &
  - &
  0.1253 &
  0.5576 &
  0.2540 &
  0.4597 &
  - &
  9.2573 &
  8.5660 &
  19.2325 &
  0.7438 &
  - &
  727.0779 &
  2.1793 &
  355.6681 &
  8.3168 &
  - \\
*top5 ensemble &
  - &
  - &
  0.8028 &
  0.6204 &
  0.6486 &
  0.6445 &
  - &
  0.5550 &
  9.4040 &
  1.8250 &
  0.5912 &
  - &
  2.2867 &
  1.0449 &
  1.2098 &
  3.0169 &
  - \\
\bottomrule
\end{tabular}%
}
\label{tab:raw_results}
\end{table}
\end{landscape}

\begin{landscape}

\begin{table}[h]
\centering
\caption{Alternative metrics, pathology removal ablation, and classification ablation. The left block reports additional voxel-level (NSD, VD) and lesion-level (\ac{fnv}, \ac{fpv}, CC-DSC, PQ, SQ, F1, F1\textsubscript{global}) metrics as weighted averages across the four test conditions, alongside \ac{dsc} computed over all samples (DSC2) and lesion-positive cases only (\ac{dsc}). The classification columns report the number of correctly identified lesion-positive (TP, n=156) and lesion-negative (TN, n=44) cases. Bold values indicate best performance per column. The pathology ablation columns show \ac{dsc} and \ac{fpv} after removing approximately 50 voxels covering the lacrimal gland region from each \ac{psma}\textsubscript{UKT} case, with the difference to the original score in parentheses.}
\def\arraystretch{1.2} 
\resizebox{1.2\textwidth}{!}{%

\begin{tabular}{llllllllllll|ll|ll}
\toprule
Algorithm &
  \multicolumn{11}{l|}{Metrics} &
  \multicolumn{2}{l|}{Classification} &
  \multicolumn{2}{l}{Pathology ablation \uktpsma} \\
\textbf{} &
  DSC &
  FNV &
  FPV &
  DSC2 &
  NSD &
  VD &
  CC-DSC &
  PQ &
  SQ &
  F1 &
  F1\textsubscript{global} &
  TP &
  TN &
  DSC &
  FPV \\ \hline
LesionTracer A &
  \textbf{0.6616} &
  \textbf{3.1842} &
  2.7849 &
  0.5709 &
  \textbf{0.7721} &
  0.5983 &
  0.5705 &
  0.4250 &
  0.5869 &
  0.6873 &
  0.7380 &
  \textbf{152} &
  12 &
  0.62 (0.05) &
  4.31 (-1.00) \\
IKIM B &
  0.6544 &
  3.7780 &
  2.3773 &
  0.6590 &
  0.7466 &
  0.6691 &
  0.5531 &
  0.4244 &
  0.5641 &
  0.6800 &
  0.7460 &
  147 &
  31 &
  \textbf{0.66 (0.00)} &
  \textbf{1.13 (-0.01)} \\
IKIM A &
  0.6496 &
  3.2451 &
  3.0438 &
  0.6300 &
  0.7390 &
  0.6492 &
  0.5609 &
  0.4165 &
  0.5745 &
  0.6606 &
  0.7356 &
  148 &
  26 &
  0.62 (0.01) &
  3.00 (-0.10) \\
StockholmTrio &
  0.6448 &
  5.4290 &
  2.1303 &
  0.6354 &
  0.7422 &
  0.6429 &
  0.5707 &
  0.4364 &
  0.5836 &
  \textbf{0.6944} &
  0.7539 &
  149 &
  28 &
  0.66 (0.03) &
  3.57 (-0.32) \\
HussainAlasmawi B &
  0.6442 &
  4.9261 &
  4.1112 &
  0.5765 &
  0.7368 &
  0.6171 &
  \textbf{0.5708} &
  0.4320 &
  \textbf{0.6005} &
  0.6762 &
  0.7464 &
  150 &
  17 &
  0.61 (0.05) &
  6.00 (-0.79) \\
HussainAlasmawi A &
  0.6438 &
  3.4792 &
  3.2942 &
  0.6176 &
  0.7367 &
  0.6417 &
  0.5610 &
  0.4305 &
  0.5861 &
  0.6720 &
  0.7382 &
  148 &
  24 &
  0.62 (0.03) &
  4.68 (-0.54) \\
AiraMatrix A &
  0.6421 &
  7.7111 &
  \textbf{1.2254} &
  \textbf{0.6740} &
  0.7419 &
  \textbf{0.6705} &
  0.5467 &
  \textbf{0.4455} &
  0.5706 &
  0.6882 &
  \textbf{0.7540} &
  144 &
  \textbf{35} &
  0.66 (0.00) &
  1.70 (-0.03) \\
QuantIF2 &
  0.6269 &
  5.4083 &
  3.1667 &
  0.6484 &
  0.7205 &
  0.6496 &
  0.5198 &
  0.4073 &
  0.5623 &
  0.6539 &
  0.7319 &
  145 &
  33 &
  0.62 (0.00) &
  3.51 (-0.03) \\
QuantIF A &
  0.6236 &
  5.3864 &
  3.9100 &
  0.6353 &
  0.7189 &
  0.6496 &
  0.5207 &
  0.4024 &
  0.5612 &
  0.6549 &
  0.7245 &
  147 &
  31 &
  0.61 (0.00) &
  4.19 (-0.07) \\
UIH-CRI-SIL A &
  0.6041 &
  3.4121 &
  4.6848 &
  0.5369 &
  0.7179 &
  0.5717 &
  0.5161 &
  0.3493 &
  0.5200 &
  0.6190 &
  0.6883 &
  153 &
  15 &
  0.43 (0.02) &
  11.42 (-1.03) \\
Lennonlychan A &
  0.5954 &
  6.3027 &
  2.5280 &
  0.5555 &
  0.7062 &
  0.6165 &
  0.5139 &
  0.3601 &
  0.5285 &
  0.6127 &
  0.7140 &
  145 &
  19 &
  0.54 (0.06) &
  4.41 (-1.50) \\
Shadab &
  0.5941 &
  5.7322 &
  3.1350 &
  0.4976 &
  0.6862 &
  0.5583 &
  0.5145 &
  0.3339 &
  0.5582 &
  0.5429 &
  0.6633 &
  147 &
  8 &
  0.55 (0.10) &
  4.13 (-2.96) \\
ZeroSugar A &
  0.5917 &
  7.3396 &
  2.2703 &
  0.5433 &
  0.7040 &
  0.6072 &
  0.5030 &
  0.3623 &
  0.5556 &
  0.5941 &
  0.6884 &
  147 &
  17 &
  0.54 (0.11) &
  3.22 (-3.09) \\
BAMF A &
  0.5902 &
  7.4716 &
  1.9465 &
  0.5903 &
  0.6924 &
  0.6161 &
  0.4842 &
  0.3651 &
  0.5188 &
  0.5981 &
  0.7252 &
  140 &
  26 &
  0.52 (0.05) &
  1.26 (-1.00) \\
BAMF B &
  0.5900 &
  6.0844 &
  2.4771 &
  0.5889 &
  0.6918 &
  0.6323 &
  0.4819 &
  0.3572 &
  0.5292 &
  0.5961 &
  0.7012 &
  143 &
  26 &
  0.51 (0.07) &
  1.88 (-1.52) \\
Lennonlychan B &
  0.5894 &
  7.4526 &
  3.2034 &
  0.5261 &
  0.7077 &
  0.6006 &
  0.5121 &
  0.3337 &
  0.5207 &
  0.5798 &
  0.6869 &
  147 &
  14 &
  0.54 (0.09) &
  3.32 (-2.20) \\
ZeroSugar B &
  0.5849 &
  8.6258 &
  2.0972 &
  0.5697 &
  0.7028 &
  0.6182 &
  0.5000 &
  0.3674 &
  0.5461 &
  0.6068 &
  0.7064 &
  146 &
  23 &
  0.56 (0.10) &
  2.65 (-2.56) \\
*datacentric baseline &
  0.5816 &
  8.2107 &
  2.2499 &
  0.5405 &
  0.6930 &
  0.5920 &
  0.4893 &
  0.3423 &
  0.5343 &
  0.5745 &
  0.6822 &
  148 &
  18 &
  0.55 (0.11) &
  2.94 (-2.82) \\
WukongRT &
  0.5504 &
  5.8178 &
  8.1048 &
  0.4790 &
  0.6434 &
  0.5409 &
  0.4769 &
  0.3210 &
  0.5332 &
  0.5436 &
  0.6719 &
  146 &
  11 &
  0.39 (0.05) &
  16.27 (-2.51) \\
HKURad &
  0.5362 &
  7.0515 &
  5.5265 &
  0.5039 &
  0.6370 &
  0.5550 &
  0.4463 &
  0.3293 &
  0.5109 &
  0.5722 &
  0.6831 &
  144 &
  18 &
  0.44 (0.09) &
  7.62 (-5.54) \\
LesionTracer B &
  0.5101 &
  14.2950 &
  3.5821 &
  0.4947 &
  0.6139 &
  0.5402 &
  0.4332 &
  0.3112 &
  0.5146 &
  0.5420 &
  0.6587 &
  144 &
  19 &
  0.46 (0.10) &
  6.90 (-3.73) \\
DING1122 &
  0.5000 &
  7.0791 &
  35.8148 &
  0.4809 &
  0.5905 &
  0.5013 &
  0.4490 &
  0.3618 &
  0.5552 &
  0.5734 &
  0.6851 &
  146 &
  20 &
  0.13 (0.02) &
  88.10 (-39.05) \\
AiraMatrix B &
  0.4957 &
  8.4701 &
  13.0775 &
  0.4563 &
  0.5922 &
  0.5122 &
  0.4140 &
  0.2751 &
  0.4805 &
  0.4865 &
  0.6060 &
  142 &
  16 &
  0.25 (0.00) &
  35.63 (-1.63) \\
UIH-CRI-SIL &
  0.4649 &
  9.4697 &
  11.6174 &
  0.3750 &
  0.5400 &
  0.4693 &
  0.3837 &
  0.2084 &
  0.4760 &
  0.3897 &
  0.4383 &
  145 &
  5 &
  0.32 (0.03) &
  26.58 (-4.06) \\
*nnunet baseline &
  0.4523 &
  7.3355 &
  31.5028 &
  0.3511 &
  0.5242 &
  0.4130 &
  0.4261 &
  0.2792 &
  0.5259 &
  0.4812 &
  0.6199 &
  148 &
  2 &
  0.19 (0.04) &
  63.99 (-38.66) \\
Maxsh A &
  0.4398 &
  10.5799 &
  242.5130 &
  0.3632 &
  0.5065 &
  0.4180 &
  0.4081 &
  0.2951 &
  0.5471 &
  0.4574 &
  0.5502 &
  143 &
  9 &
  0.03 (0.00) &
  876.16 (-48.15) \\
Maxsh B &
  0.4274 &
  13.0200 &
  239.9951 &
  0.3656 &
  0.4931 &
  0.4141 &
  0.3980 &
  0.2926 &
  0.5298 &
  0.4561 &
  0.5511 &
  142 &
  12 &
  0.03 (0.00) &
  876.16 (-48.15) \\
Shrajanbhandary &
  0.3567 &
  24.6515 &
  5499.5501 &
  0.3324 &
  0.4530 &
  0.3995 &
  0.2956 &
  0.1966 &
  0.3357 &
  0.3396 &
  0.5509 &
  107 &
  12 &
  0.38 (0.02) &
  20.28 (-4.48) \\
TUM-ibbm &
  0.2601 &
  11.2489 &
  197.2717 &
  0.2007 &
  0.2977 &
  0.2794 &
  0.2388 &
  0.1711 &
  0.4459 &
  0.3166 &
  0.4281 &
  136 &
  0 &
  0.20 (0.01) &
  79.66 (-10.26) \\ \bottomrule
\end{tabular}

}
\label{tab:alternative_metrics}
\end{table}
\end{landscape}

\onecolumn

\begin{landscape}
\vspace*{\fill} 
\begin{figure}[!ht]
\centering
\includegraphics[width=\columnwidth]{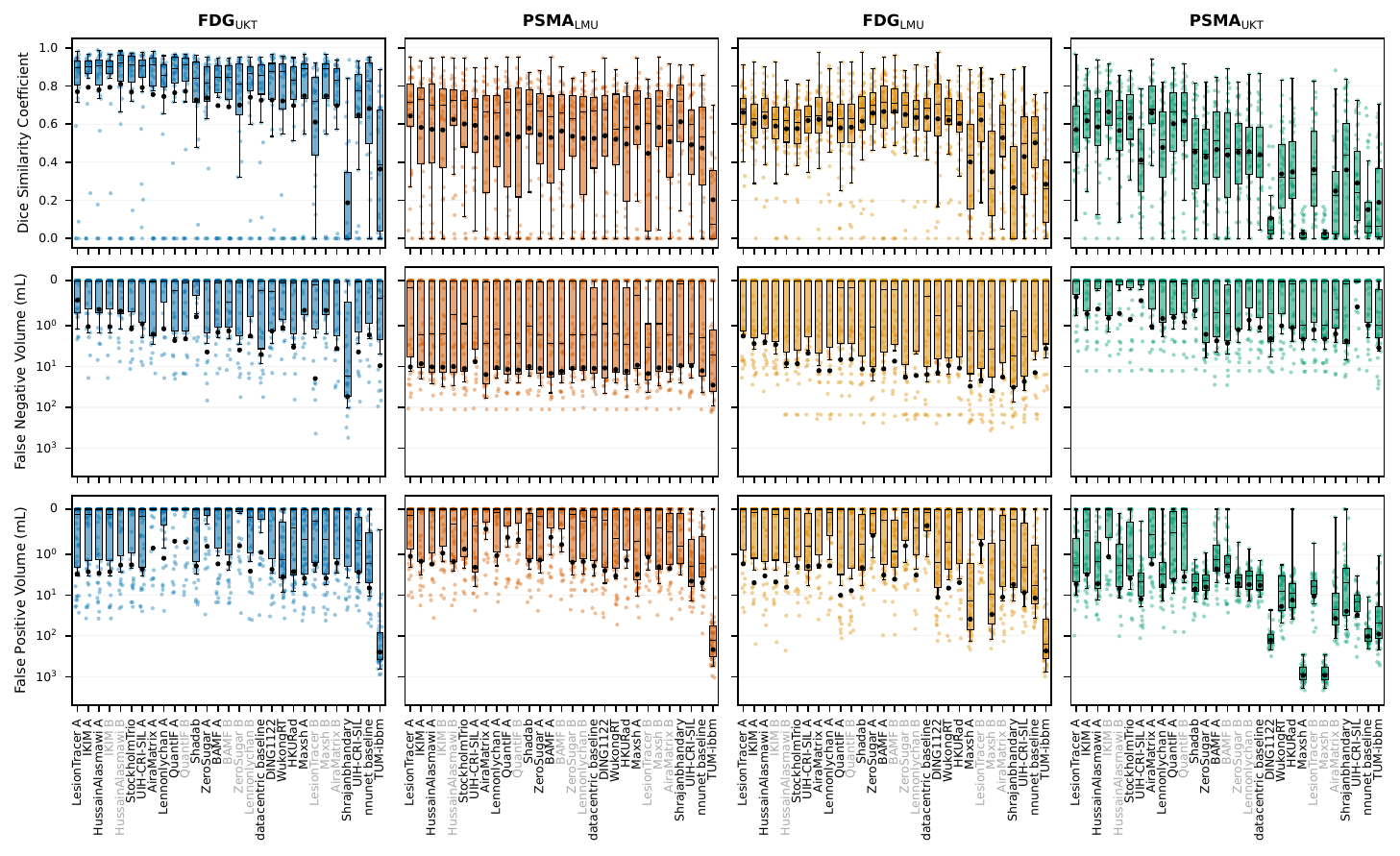}
\caption{Per-algorithm performance stratified by dataset. Each column corresponds to one dataset (\uktfdg, \lmupsma, \lmufdg, \uktpsma), and each row to one metric (Dice Similarity Coefficient, False Negative Volume, False Positive Volume). Box plots show the distribution across test cases for every submitted algorithm. Note the inverted logarithmic scale for \ac{fnv} and \ac{fpv}, where higher boxes indicate lower volumetric errors. This figure complements the aggregated view in Figure~\ref{fig_overview_results} and the averaged results in Table~\ref{tab:raw_results} by making per-case variability and outlier patterns visible for each tracer–center combination individually.}\label{fig:individual_metrics}
\end{figure}
\vspace*{\fill} 
\end{landscape}

\begin{landscape}
\vspace*{\fill} 
\begin{figure}[!ht]
\centering
\includegraphics[width=\columnwidth]{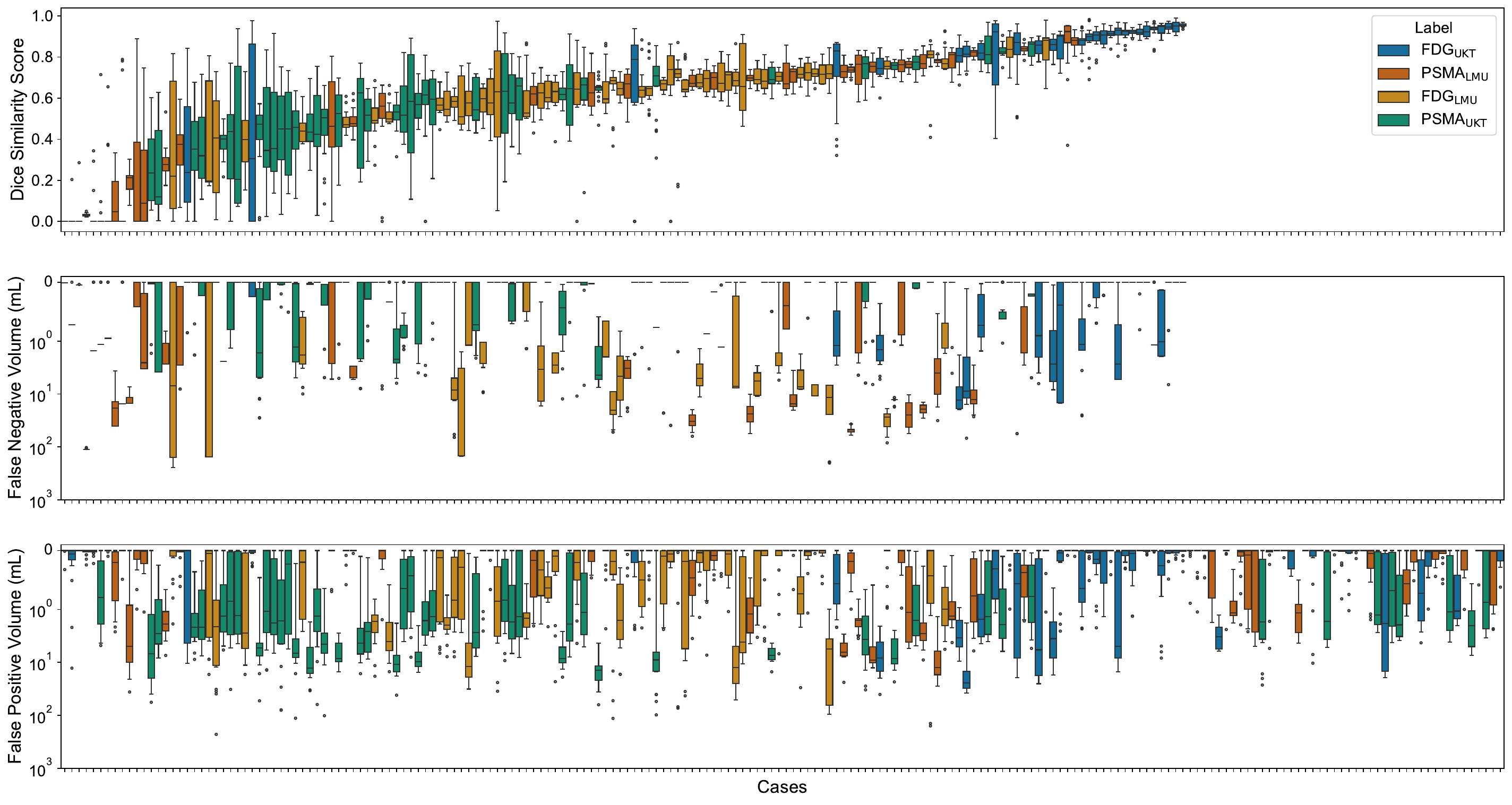}
\caption{Per-case performance distribution across the top-18 algorithms. Cases are ordered by average \ac{dsc} along the $x$-axis. Rows show the Dice Similarity Coefficient (top), false-negative volume in mL (middle), and false-positive volume in mL (bottom); volume axes use a logarithmic scale. Box plots are color-coded by dataset condition. A small cluster of near-zero \ac{dsc} cases is visible on the left, corresponding to patients with only a single small lesion. The right side is dominated by \uktfdg\ samples with high median scores and narrow interquartile ranges, while \lmufdg\ patients populate the mid-range and \lmupsma\ is scattered across the full spectrum. \uktpsma\ consistently shows larger variance and lower \ac{dsc}.}
\label{fig:sample_dist}
\end{figure}
\vspace*{\fill} 
\end{landscape}

\begin{figure*}[!ht]
\centering
\includegraphics[width=0.99\textwidth]{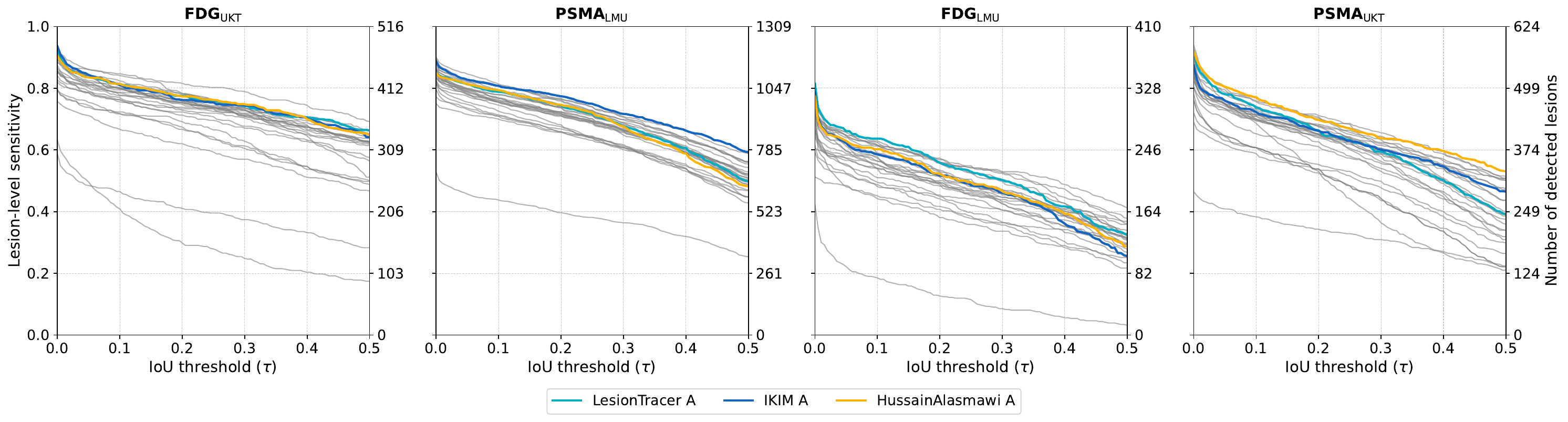}
\caption{Per-dataset lesion detection sensitivity as a function of the IoU threshold $\tau$, stratified by center and tracer. The left $y$-axis shows lesion-level sensitivity; the right $y$-axis shows the absolute number of detected lesions. Three of the four conditions median start above 0.84 at the one-voxel criterion, whereas \lmufdg\ begins at approximately 0.74 and exhibits a steeper decline across the threshold range. Top 3 teams are highlighted; gray lines denote remaining submissions.}
\label{fig:detection_deciles_per_ds}
\end{figure*}

\begin{figure*}[!ht]
\centering
\includegraphics[width=0.99\textwidth]{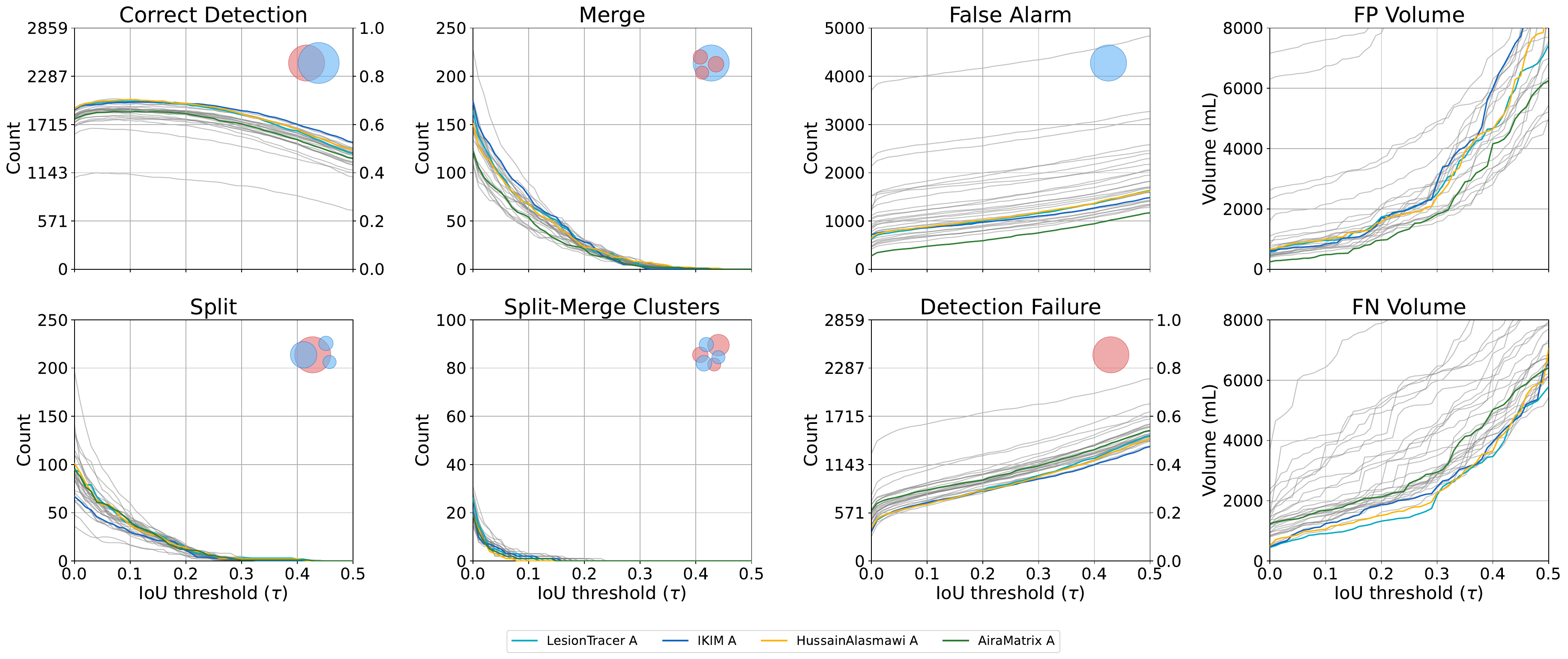}
\caption{Detection error decomposition as a function of the IoU threshold $\tau$. The six panels on the left show the count of each error type: correct detections (CD), merges (M), false alarms (FA), splits (S), split-merge clusters (SM), and detection failures (DF). Bubble diagrams illustrate the association pattern between ground-truth labels (red) and predictions (blue). The two right-hand panels show the cumulative false-positive and false-negative volume in mL. At the one-voxel criterion, a substantial number of merge, split, and split-merge associations exist; as $\tau$ increases, these cluster associations decay toward zero, resolving into correct detections, detection failures, and false alarms. Beyond $\tau \approx 0.3$, FP and FN volumes grow more steeply as nearly all cluster associations have vanished. Top 4 teams are highlighted; gray lines denote remaining submissions.}
\label{fig:nascimiento}
\end{figure*}

\begin{figure*}[!ht]
\centering
\includegraphics[width=0.99\textwidth]{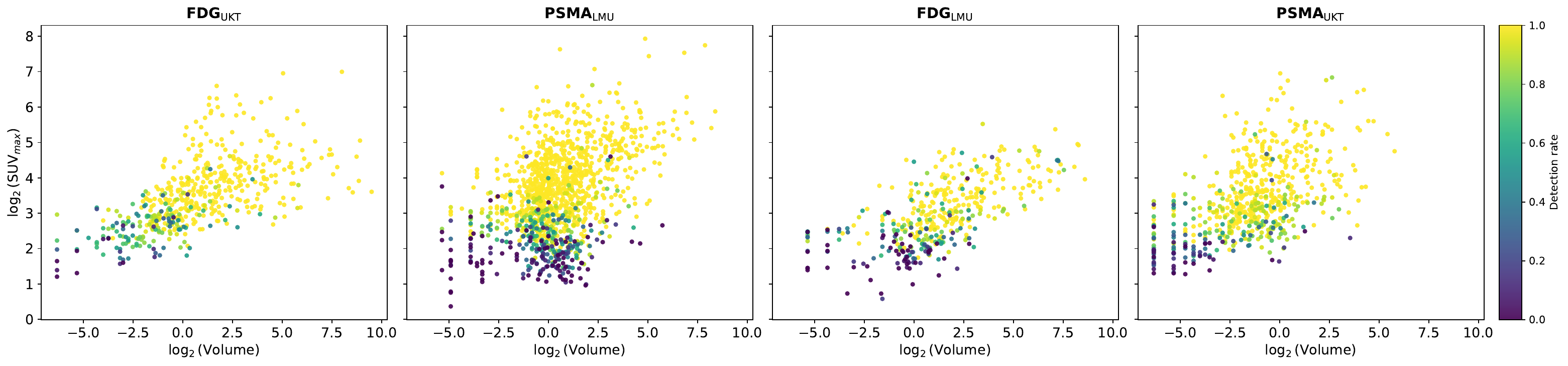}
\caption{Per-dataset lesion detection probability across top-18 algorithms depending on lesion volume and SUV\textsubscript{max}. Consistently, very small lesions (roughly \textless~0.2\,mL) and low-uptake lesions (SUV\textsubscript{max} roughly \textless~4) are missed across all four test conditions, as reflected by the dark purple points concentrated in the lower-left of each panel. Detection probability increases jointly with both volume and uptake.}
\label{fig:suv_uncert}
\end{figure*}
\twocolumn

\bibliographystyle{elsarticle-harv}
\bibliography{autopet3}

\end{document}